\documentclass[conference]{IEEEtran}

\AtBeginDocument{%
  }

\usepackage{ulem}
\usepackage{makecell}
\usepackage{multirow}
\usepackage[utf8]{inputenc} 
\usepackage{threeparttable}
\usepackage{url}            
\usepackage{booktabs}       
\usepackage{amsfonts}       
\usepackage{nicefrac}       
\usepackage{microtype}      
\usepackage{subfig}
\usepackage{soul}
\usepackage{colortbl}
\usepackage{booktabs}

\usepackage{tabularx}
\usepackage{makecell}
\usepackage{wrapfig}
\usepackage{mathtools}
\usepackage{graphicx}
\usepackage{bbm}
\usepackage[linesnumbered,ruled,vlined]{algorithm2e}

\usepackage{array}
\usepackage{caption}
\usepackage{textcomp}
\usepackage{amsmath}
\usepackage{balance}
\usepackage[ruled,vlined]{algorithm2e}
\SetKwInput{KwInput}{Input}
\SetKwInput{KwOutput}{Output}
\SetKwComment{Comment}{\# }{}
\usepackage{xcolor}
\usepackage{array}
\usepackage{float}
\usepackage{ulem} 
\usepackage{hyperref}

\usepackage{wrapfig}

\newtheorem{definition}{Definition}

\IEEEoverridecommandlockouts

\newcommand{\model}{\textsc{FastFT}}
\DeclareMathOperator*{\argmax}{argmax} 

\begin{document}

\title{\model: Accelerating Reinforced Feature Transformation via Advanced Exploration Strategies}
\author{\IEEEauthorblockN{
Tianqi He$^{1,3,\S}$, Xiaohan Huang$^{1,2,\S}$, Yi Du$^1$, Qingqing Long$^{1,2}$, Ziyue Qiao$^{4,*}$, Min Wu$^{5}$\\ Yanjie Fu$^{6}$, Yuanchun Zhou$^{1,2}$, Meng Xiao$^{1,*}$\thanks{$^*$Corresponding authors: Meng Xiao (shaow@cnic.cn) and Ziyue Qiao (zyqiao@gbu.edu.cn). $^\S$Tianqi He and Xiaohan Huang contributed equally to this work. This work is partially supported by the National Natural Science Foundation of China (No.92470204, No.62406056), Beijing Natural Science Foundation (No.4254089), and China Postdoctoral Science Foundation (No.2023M743565, No.GZC20232736).}}
\IEEEauthorblockA{\textit{$^1$Computer Network Information Center, Chinese Academy of Sciences, China.} \\
\textit{$^2$University of Chinese Academy of Sciences, China.} \\
\textit{$^3$School of Advanced Interdisciplinary Sciences, University of Chinese Academy of Sciences, China.} \\
\textit{$^4$School of Computing and Information Technology, Great Bay University, China.} \\
\textit{$^5$Agency for Science, Technology and Research (A*STAR), Singapore.} 
\textit{$^6$Arizona State University, USA.}} 
}

\maketitle
\begin{abstract}
Feature Transformation is crucial for classic machine learning that aims to generate feature combinations to enhance the performance of downstream tasks from a data-centric perspective. 
Current methodologies, such as manual expert-driven processes, iterative-feedback techniques, and exploration-generative tactics, have shown promise in automating such data engineering workflow by minimizing human involvement. 
However, three challenges remain in those frameworks: 
(1) It predominantly depends on downstream task performance metrics, as assessment is time-consuming, especially for large datasets.
(2) The diversity of feature combinations will hardly be guaranteed after random exploration ends.
(3) Rare significant transformations lead to sparse valuable feedback that hinders the learning processes or leads to less effective results. 
In response to these challenges, we introduce \model, an innovative framework that leverages a trio of advanced strategies.
We first decouple the feature transformation evaluation from the outcomes of the generated datasets via the performance predictor. 
To address the issue of reward sparsity, we developed a method to evaluate the novelty of generated transformation sequences. 
Incorporating this novelty into the reward function accelerates the model's exploration of effective transformations, thereby improving the search productivity. 
Additionally, we combine novelty and performance to create a prioritized memory buffer, ensuring that essential experiences are effectively revisited during exploration.
Our extensive experimental evaluations validate the performance, efficiency, and traceability of our proposed framework, showcasing its superiority in handling complex feature transformation tasks\footnote{
The code and data are publicly accessible via ~\href{https://github.com/coco11563/FASTFT-Accelerating-Reinforced-Feature-Transformation-via-Advanced-Exploration-Strategies}{Github}.}. 
\end{abstract}



\section{Introduction}

Feature Transformation (FT) plays a pivotal role in enhancing the performance of downstream machine learning models~\cite{zhang2023openfe,icde24feataug,ionescu2024autofeat} by generating high-quality datasets rich in information with mathematical transformations and integrating features from a data-centric approach~\cite{zha2023data,dcai}. 
Those generated features can significantly improve the precision of downstream tasks such as classification, regression, and detection tasks~\cite{domingos2012few} for machine learning models~\cite{hancock2020survey, borisov2022deep}. 
Traditionally, feature transformation has relied heavily on the extensive knowledge and significant manpower of domain experts~\cite{bengio2013representation,conrad2022benchmarking}, making the process both time-consuming and inefficient. 
To address these issues, current research primarily focuses on automating this pipeline through advanced technologies such as evolutionary algorithms~\cite{tran2016genetic,luo2019autocross,liu2024interpretable}, reinforcement learning~\cite{kdd2022}, and generative models~\cite{zhu2022difer}. 
Although existing automated methods have decoupled the dataset generation pipeline from expert evaluations, they still rely on automated assessment approaches (such as evaluating the performance of the generated datasets on downstream tasks) to gauge the quality of the datasets produced~\cite{wang2024reinforcement}. 
Consequently, the time consumption of these partially automated approaches remains dependent on the scale of the dataset~\cite{hollmann2024large}.
The work~\cite{huang2024enhancing} highlights that evaluations of downstream tasks account for up to 80\% of the total runtime. 
This scalability limitation poses a significant runtime bottleneck in feature transformation tasks, severely impacting their efficiency and practical applicability.
Furthermore, meaningful feature-crossing is rare within the infinite search space. 
Some studies~\cite{xiao2022traceable,xiao2024traceable} initially enhance the diversity of feature transformations explored using random search methods. 
However, as trained strategies begin to dominate the entire framework, most transformations do not produce significant effects. 
This results in highly sparse rewards within the training framework, making the model's training unstable and prone to converging on local optima.

\begin{figure}[!t]
\centering
\includegraphics[width=\linewidth]{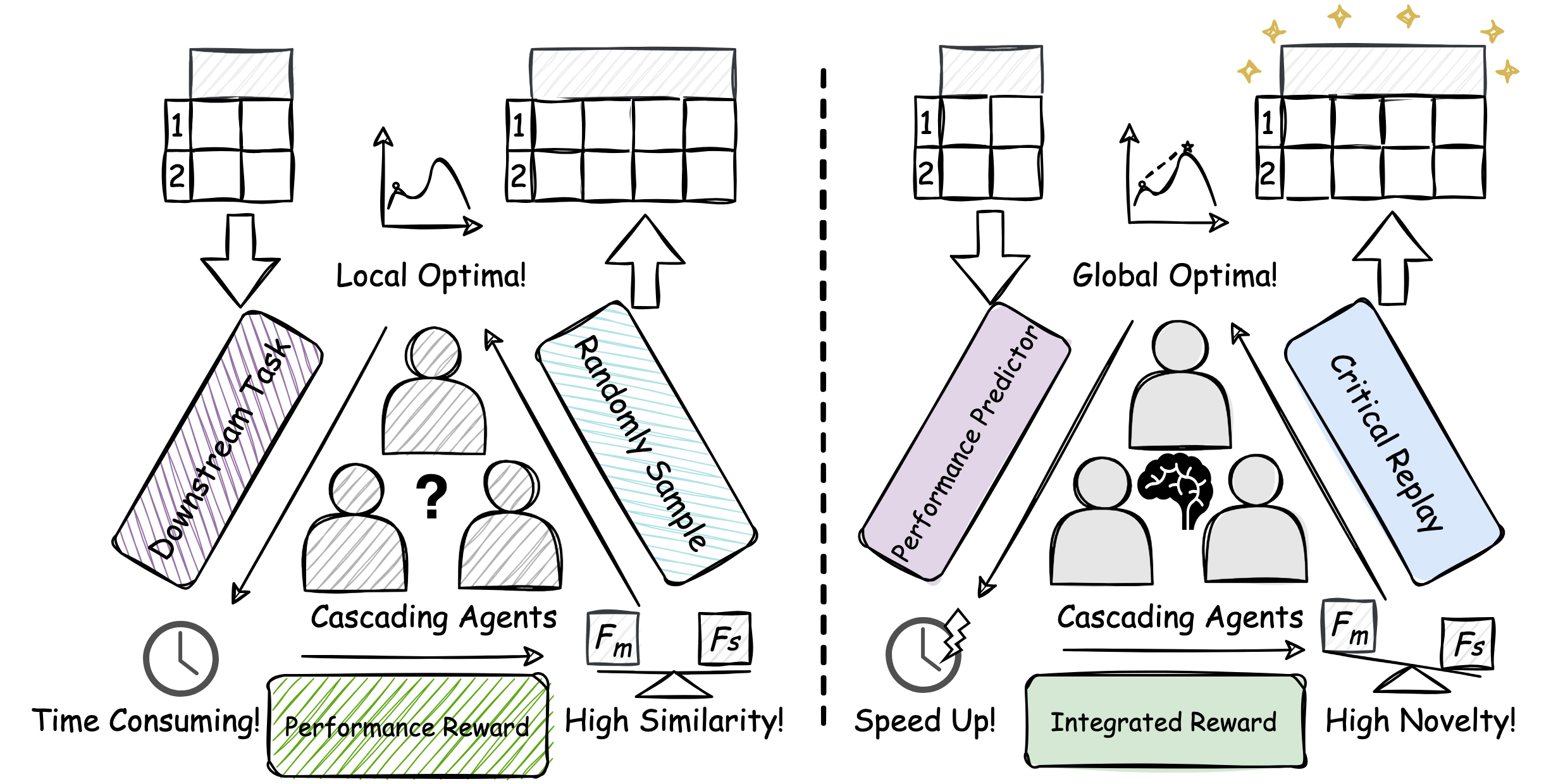}
    \caption{The illustration of our motivation. 
    The left figure shows the challenges of existing feature transformation approaches.
    The right figure explains our framework can speed up the iteration and improve the training stability.}
    \vspace{-0.5cm}
    \label{fig:motivation}
\end{figure}

From the discussion above, we summarize three significant challenges in the reinforcement FT task (as depicted in the left panel of Figure~\ref{fig:motivation}): \textbf{(C1: Runtime Bottleneck)} \uline{Evaluation of the evolving transformation strategy is based on the time-consuming feedback from downstream tasks. }
This excessive reliance on performance metrics from downstream tasks not only extends the time required to validate transformations but also reduces the process's agility~\cite{ying2024unsupervised}, potentially delaying insights that could lead to more immediate improvements in model performance.
\textbf{(C2: Local Optimal)} \uline{The diversity of feature cross will hardly be guaranteed after random exploration ends.} 
As the exploration phase transitions to more strategy-driven approaches, there is a significant risk that the automated systems will converge prematurely on a limited set of feature combinations~\cite{rnd,yang2024exploration}, potentially overlooking novel or unconventional interactions that could offer substantial benefits to the model's performance~\cite{antirnd}.
\textbf{(C3: Rare Significance)} \uline{Meaningful and impactful transformations infrequently occur.}
Faced with the expansive search space for feature transformations, our systems frequently encounter a scarcity of impactful outcomes, leading to a slow training process characterized by sparse rewards that challenge the efficacy of learning.


\textbf{Our Perspectives and Insights:}  
As illustrated in the right panel of Figure~\ref{fig:motivation}, we delve into the limitations of current feature transformation frameworks and introduce our innovative perspectives to address these challenges.
\textbf{(1) Replace poor-scalable feedback with an adaptive adopted empirical evaluation.} 
We employ an empirical evaluation method that provides faster and more detailed feedback on the efficiency of each transformation step, thus significantly alleviating the dependence on the feedback of poorly scalable and time-intensive downstream tasks.
\textbf{(2) Encourage Novel Feature Combination as Reward to Overcome Sparse Reward Issue.} We introduce a novelty estimation technique. This method assigns rewards based on the novelty and potential utility of the transformations, encouraging the exploration of new and potentially more effective feature transformations. This strategy ensures a more stable and continuous learning process by providing frequent and meaningful feedback.
\textbf{(3) Replay Critical Transformation for an Effective Optimization Pipeline.} Recognizing the importance of meaningful transformations, our framework incorporates a mechanism to replay critical transformations. 
By prioritizing and revisiting transformative steps that have previously shown significant impact, we can effectively tune the feature transformation process and achieve superior results.

\textbf{Summary of Framework: An Efficient Reinforced Feature Transformation Framework.} 
To capitalize on the benefits of the aforementioned perspectives, we introduce the \textit{\textbf{Fast} \textbf{F}eature \textbf{T}ransformation Framework} (\textbf{\model}).
The \model\ framework is designed to address the intrinsic challenges of feature transformation by incorporating advanced exploration strategies.
Our framework can be divided into four stages. Specifically, during the \textit{initial exploration stage}, we explore feature-feature crossing strategies and assess the generated dataset based on the performance of downstream tasks. 
Once a diverse set of transformation-score pairs has been collected, the framework transitions to the \textit{evaluation component training stage}.
In this stage, we train two components with collected data for reward estimation. 
The Performance Predictor is designed to assess the transformation sequence, and the Novelty Estimator aims to measure the distinction between the generated sequence and the collected sequence. 
These components will replace the feedback from the downstream task by estimating the reward. 
The framework then moves to \textit{the effective exploration stage}.
In this stage, downstream task feedback will complementarily assess only the transformation sequences that exhibit a high effect on downstream tasks and high novelty, thus reducing the whole system run-time bottleneck. 
These memories will also be preserved in a prioritized memory buffer and then used to optimize cascading agents. 
As the model explores more unencountered transformations, the framework repeatedly progresses between \textit{the fine-tuning stage} and \textit{the effective exploration stage}. 
In this stage, all collected high-quality transformations in the memory buffer are used to fine-tune the two evaluation components. 

We conduct extensive experiments and case studies to validate the effectiveness of each technical component.
The qualitative and quantitative analysis results show a significant improvement in learning efficiency, performance, traceability, quality of the generated features, and time consumption.

\section{Problem Formulation and Important Definition}

\subsection{Important Definitions}

\begin{definition}
\textbf{Operation Set.}
To generate a new feature, we perform a mathematical operation to one or two existing features, e.g., $f_1+f_2$ as a new feature. The set of operations, denoted by $\mathcal{O}$, is categorized into unary operations and binary operations. Unary operations, like ``square'', ``exp'', and ``log'', are applied to a single feature. Binary operations, such as ``plus'', ``multiply'', ``divide'', are applied to two features. 
\end{definition}

\begin{definition}
\textbf{Feature Set.} 
We denote a dataset as $\mathcal{D} = <\mathcal{F},y>$, where $\mathcal{F} = \{F_1, F_2, \ldots\}$ are original or generated feature (column) set. $F_i$ is the $i$-th feature. The label for a sample (instance, row) is denoted by $y$. Throughout a feature transformation process, labels remain unchanged, but features are transformed over time.  
\end{definition}

\begin{figure}[!h]
\centering
\vspace{-0.1cm}
\flushleft
\includegraphics[width = 0.80\linewidth]{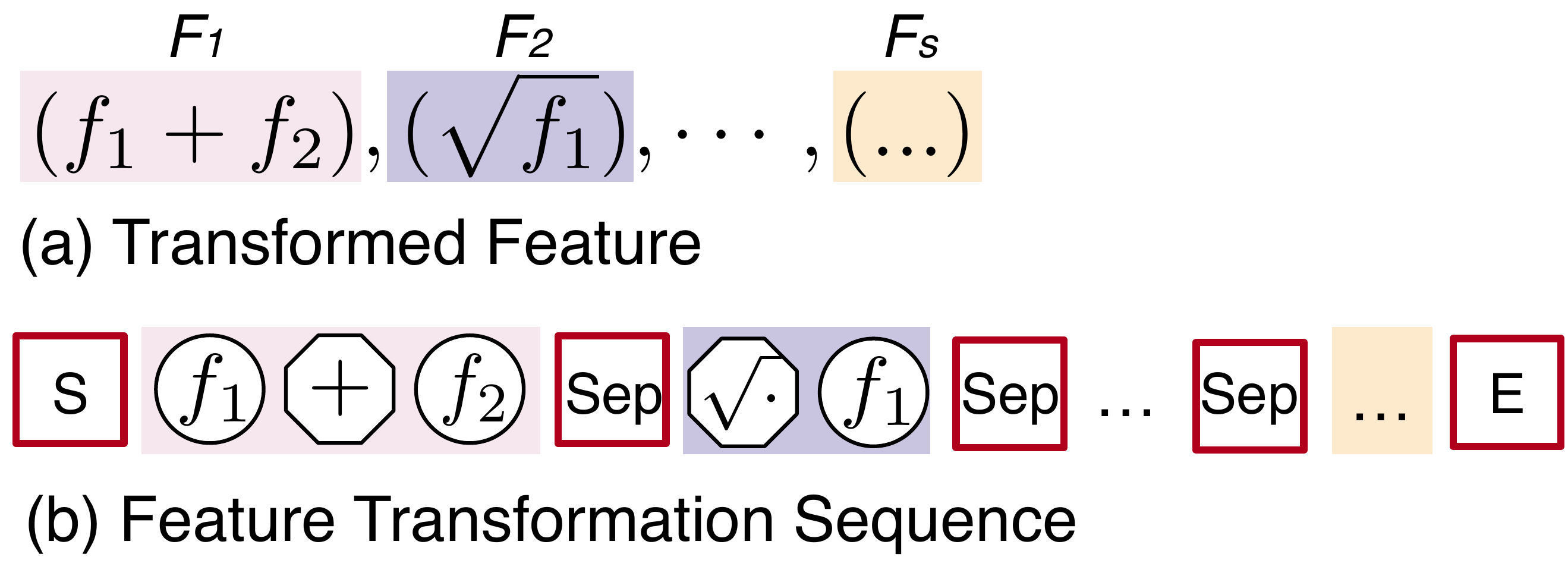}
\caption{Transformed feature set and its correlated feature transformation sequence.}
\vspace{-0.3cm}
\label{ft_seq}
\end{figure}

\begin{figure*}[!t]
\centering
\includegraphics[width=\linewidth]{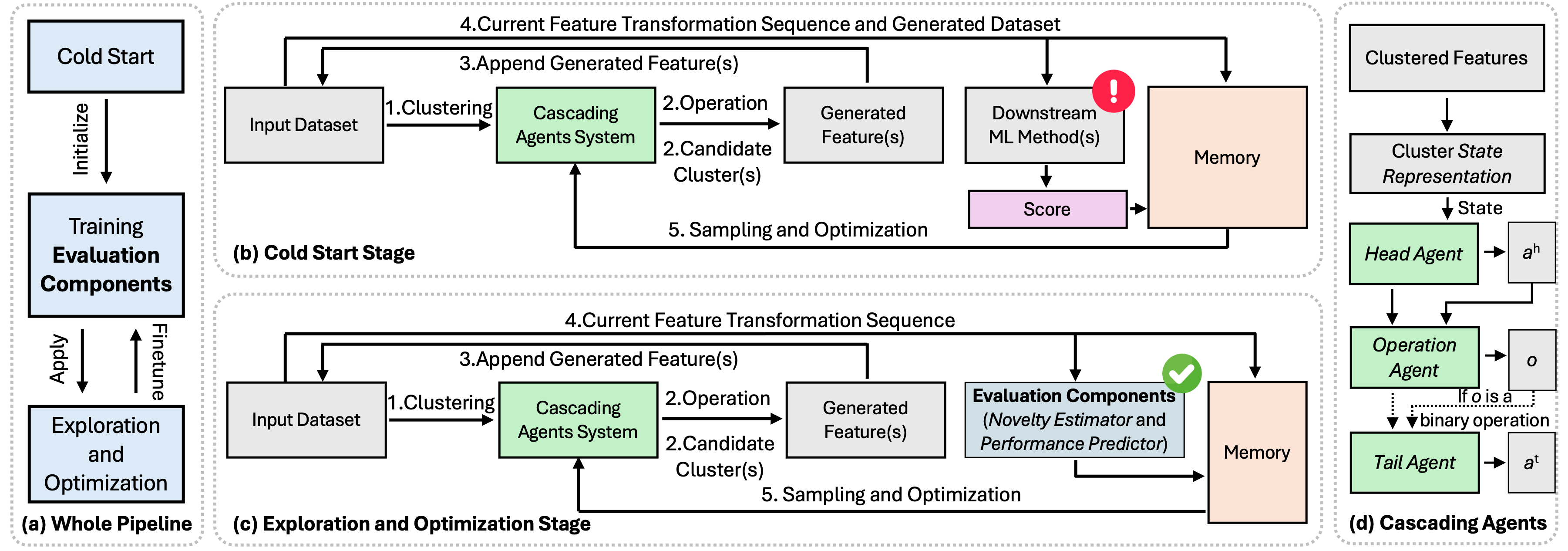}
    \caption{{\color{black} An overview of \model\ framework: (a) The illustration of the whole pipeline. (b) The cold start stage. We mark the runtime bottleneck with the exclamation point. (c) The exploration and optimization stage. By replacing the runtime bottleneck with two Evaluation Component (Novelty Estimator and Performance Predictor), the model could perform highly efficient exploration, evaluation, and optimization. (d) The cascading multi-agent system.}}
    \vspace{-0.5cm}
    \label{fig:main}
\end{figure*}

\begin{definition}
\textbf{Cascading Reinforcement Agents.}
We develop cascading reinforcement learning agents to perform unary or binary mathematical operations on single or paired candidate features, denoted as $\pi(\cdot)$. 
This configuration consists of three agents: two select features and the third select an operation. 
The three agents select features or operations, thus changing the environment states sequentially. 
The cascading design allows an agent to pass updated state representations to the next agent, enabling the subsequent agent to perceive environmental changes and make more accurate decisions.
\end{definition}

\begin{definition}
\textbf{Feature Transformation Sequence.} A transformation process is represented by a sequence of feature transformation tokens: $T = [t_1, t_2, \dots]$. Figure~\ref{ft_seq} shows that a token $t_i \in T$ can be a feature, an operation, or a special token, such as a starting token, an ending token, and a separation token. 
A transformation sequence, if applied to an original dataset, can create a transformed feature set, which is denoted by $T(\mathcal{F})\rightarrow \mathcal{\hat{F}}$, where $\mathcal{\hat{F}}$ denotes the transformed feature set. 

\end{definition}
\subsection{The Feature Transformation Problem}
The feature transformation problem aims to learn an optimal and explicit feature representation space given a downstream ML task and an original dataset. Formally, assuming a dataset $D=<\mathcal{F},y>$ that consists of an initial feature set $\mathcal{F}$, and a target label set $y$, along with an operator set $\mathcal{O}$. 
For a downstream ML task $A$ (such as classification, regression, detection), let $\mathcal{F}^{*}$ be its optimal feature set, $\mathcal{A}$  be the performance metric of the task $A$. 
The target is to identify the optimal transformation sequence $T^*$ that maximizes:
\begin{equation}
\label{objective}
    T^{*} = \argmax_{T\in\mathcal{T}}( \mathcal{A}(T(\mathcal{F}),y)),
\end{equation}
where $\mathcal{T}$ is a set of all possible feature transformation sequences. Finally, the optimal feature set can be transformed from the original feature set via $T^*$, given as $T^*(\mathcal{F})\rightarrow \mathcal{F}^*$.
\section{Proposed Framework}


In this section, we present an overview, and then detail each technical component of our framework.

\subsection{Framework Overview} 
Figure~\ref{fig:main} (a) shows an overview of our proposed framework, \model{}. 
Our framework has three interactive parts: 
\textit{(1) Cascading Reinforcement Learning System} (Section~\ref{rl}).
It consists of three interconnected agents. Each agent is tasked with selecting either a head feature cluster, a mathematical operation, or a tail feature cluster. This system is the backbone of our feature transformation process.
\textit{(2) Training Evaluation Components} (Section~\ref{cold_start}).
The cascading RL system generates feature-feature crossing decisions, which are assessed based on the performance of downstream tasks in the initial exploration. 
Transformation feature sequences are input into two evaluation components: the Performance Predictor and the Novelty Estimator. The former is trained using data from the transformation sequences and their corresponding downstream task performance, while the latter is trained to remember these sequences' structural information. 
\textit{(3) Efficient Exploration, and Optimization} (Section~\ref{opt}).
This part leverages the two previously mentioned evaluation components to adaptively replace the need for downstream task evaluation, thereby decoupling the dependency on the generated dataset and improving time consumption. 
The framework will also preserve transformations into the memory buffer that yield significant performance changes or high novelty to optimize the cascading agents and periodically fine-tune the evaluation components. 


\subsection{Cascading Reinforcement Learning System} \label{rl}

Cascading reinforcement learning is a multi-stage learning process where agents make sequential decisions. As shown in Figure~\ref{fig:main} (d), each agent's action is based on the previous ones, forming a cascade-like structure. 


\noindent\textbf{Feature Clustering.} Cluster-wise feature transformation can speed up generation and exploration, improve reward feedback, and help agents learn clear policies~\cite{kdd2022}. 
We adopted an incremental clustering method to handle the evolving number of features. 
In the initial stage, all features are regarded as clusters. 
In each step, our clustering method will merge the two closest clusters of features. 
The iteration will stop when the distance between the two closest clusters reaches a certain threshold.
The distance function is given by:
\begin{equation}
    \label{fea_dis}
    dis_{ij} =
    \frac{1}{|\mathcal{C}_i|\cdot|\mathcal{C}_j|}
    \sum_{F_i\in \mathcal{C}_i}\sum_{F_j\in \mathcal{C}_j}
    \frac{|MI(F_i,y)-MI(F_j,y)|}{MI(F_i,F_j)+\varsigma},
\end{equation}
where $\mathcal{C}_i$ and $\mathcal{C}_j$ represent two clusters of features.
$MI(\cdot)$ denotes the mutual information. 
$|\mathcal{C}_i|$ and $|\mathcal{C}_j|$ indicate the number of features within $\mathcal{C}_i$ and $\mathcal{C}_j$ respectively. $F_i$ and $F_j$ are individual features in $\mathcal{C}_i$ and $\mathcal{C}_j$ respectively, where $y$ represents their label vector. 
Specifically, $|MI(F_i,y)-MI(F_j,y)|$ measures the difference in importance between $y$ and features $f_i$, $f_j$. 
$MI(F_i,F_j)$ captures the level of redundancy between $F_i$ and $F_j$. 
$\varsigma$ is a small nonzero constant used to avoid zero-division.

\begin{figure}[!t]
\centering
\includegraphics[width = \linewidth]{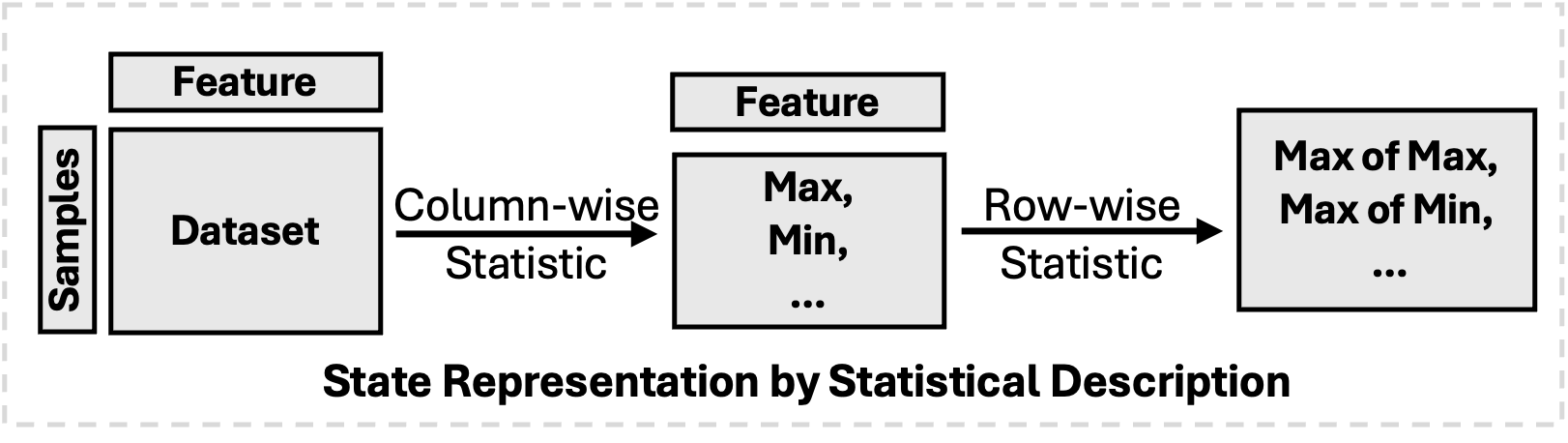}
\caption{{\color{black} The pipeline of state representation on overall or cluster of features.}}
\vspace{-0.5cm}
\label{ft_explain}
\end{figure}

\noindent\textbf{State Representation.}
We use $Rep(o)$ to denote a one-hot encoding for operation $o$ from the fixed-size operations set. 
The state representation method on the feature cluster adheres to the common configuration~\cite{xiao2024traceable}, serving as the statistical description of the column and row values (as depicted in Figure~\ref{ft_explain}). 
We use $Rep(\mathcal{C})$ and $Rep(\mathcal{\hat{F}})$ to denote the state of the clustered feature or the overall feature set. 

\noindent\textbf{Cascading Reinforcement Learning Agents.}
The cascading reinforcement learning system is developed to select a head cluster, a mathematical operation, and a tail cluster sequentially, thus constructing a function to generate a new feature.  
Formally, the feature set of the current step can be denoted as $\mathcal{\hat{F}}$. 

\noindent\textit{\uline{(1) Head Agent}}: 
The first agent aims to select the head cluster to be transformed according to the current state of each cluster. 
Specifically, the state of $i$-th feature cluster is given as $Rep(\mathcal{C}_i)$, and the overall state can be represented as $Rep(\mathcal{\hat{F}})$.
With the policy network $\pi_h(\cdot)$, the score of select $\mathcal{C}_i$ as the action can be estimated by: $s_i^h = \pi_h(Rep(\mathcal{C}_i)\oplus Rep(\mathcal{\hat{F}}))$, where $\oplus$ denotes the concatenate operation, and $a^h$ denotes the selected cluster. 

\noindent\textit{\uline{(2) Operation Agent}}:
The operation agent aims to select the mathematical operation from $\mathcal{O}$ according to the selected head feature and the overall state, defined as $o = \pi_o(Rep(a^h)\oplus Rep(\mathcal{\hat{F}}))$. Here, $a^o$ denotes the selected operation. 

\noindent\textit{\uline{(3) Tail Agent}}:
If the operation agent selects a binary operation, the tail agent will select an additional feature for the transformation process. Similar to the head agent, the policy network $\pi_t(\cdot)$ will use the state of the chosen head feature, the selected operation, the state of the overall feature set, and each potential feature as input, represented as $s_i^t = \pi_t(Rep(a^h)\oplus Rep(\mathcal{\hat{F}})\oplus Rep(a^o)\oplus Rep(\mathcal{C}_i))$. The tail cluster with the highest score is denoted as $a^t$. 
These agents collaborate and constitute a single exploration step. 

{\color{black}
\noindent\textbf{Group-wise Feature-Crossing.} 
These stages above are referred to as one exploration step. 
Depending on the selected head cluster $a_h$, operation $o$, and tail cluster $a_t$, \model\ will cross each feature and then update the transformation sequence. 
If the selected operation is binary, for each feature $a^i_h\in a_h$ and $a^j_t\in a_t$, operation $o$ is applied, resulting in the features $\{o(a^i_h,a^j_t)\}_{i,j}$, yielding a total of $|a_h| \times |a_t|$ features. 
For a unary operation, the generated feature set will remain the same size as the selected head cluster, given as $\{o(a^i_h)\}_i$.
}

\begin{figure}[!t]
\centering
\includegraphics[width = \linewidth]{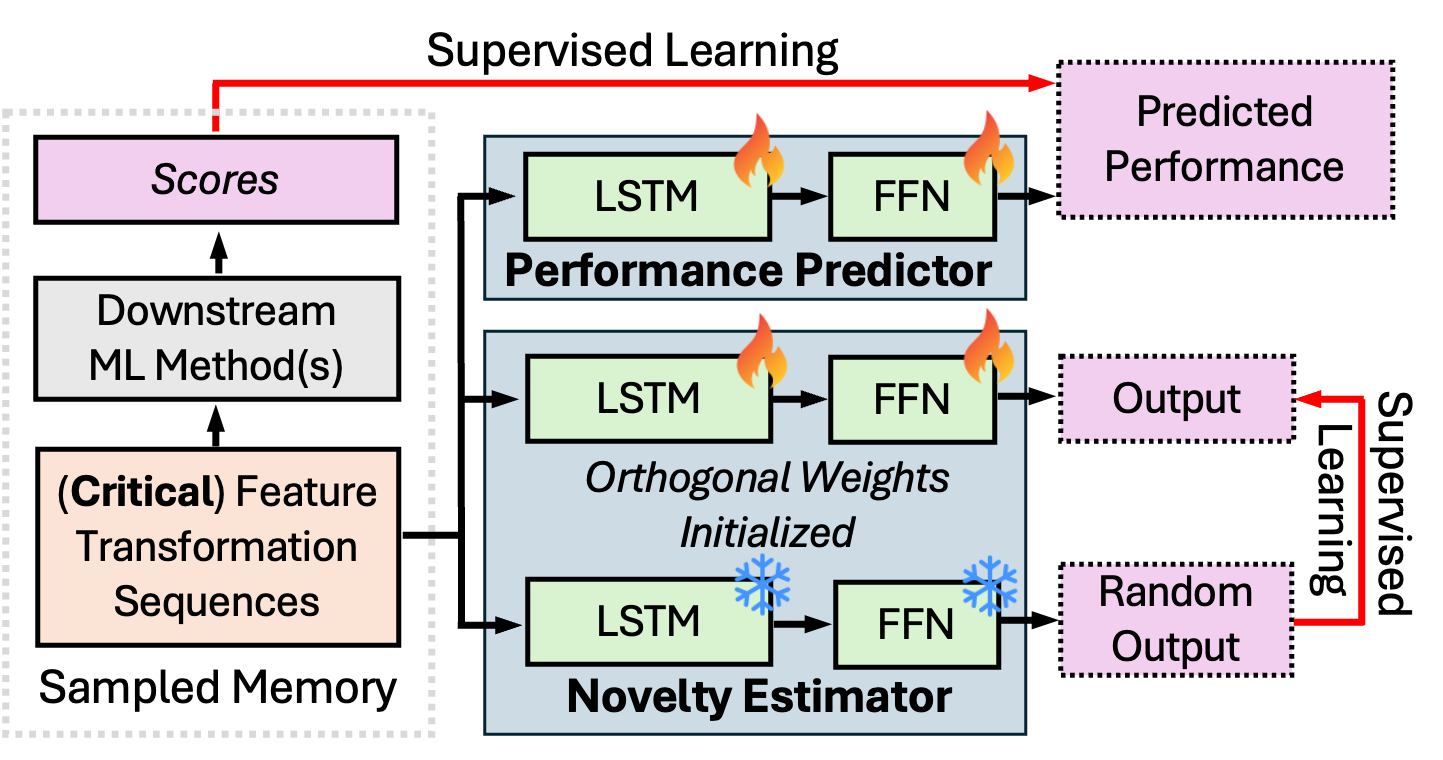}
\caption{{\color{black} The cold-start and the finetuning progress of the Evaluation Components.}}
\vspace{-0.5cm}
\label{fintune}
\end{figure}

\subsection{Evaluation Components for Fast Reward Estimation} \label{cold_start}

In this section, we introduce two key components for replacing traditional downstream task evaluation with fast-reward estimation. 

\noindent\textbf{Performance Predictor.} 
The Performance Predictor aims to reduce the time-consuming nature of downstream task evaluation by evaluating the transformation sequence using empirical task performance and sequential information. 
Specifically, the predictor takes the generated transformation sequence as input and outputs its estimated downstream performance, denoted as $\varphi(T): T \rightarrow \mathbb{R}$, where $\varphi(\cdot)$ is an LSTM~\cite{lstm} with a feedforward network. 

{\color{black}
\noindent\textbf{Novelty Estimator.} 
The Novelty Estimator consists of a fixed target network and an estimator network, both built with LSTM and feed-forward layers. 
The target network is randomly initialized and fixed, while the estimator network is orthogonally initialized~\cite{osband2018randomized, rnd} to ensure independence. 
The estimator network will be trained on the collected sequence data to minimize the prediction error between its output and the corresponding target network’s fixed output. 
As the target network is orthogonally initialized and remains frozen, this configuration leads to low prediction errors on observed features but elevated errors on unencountered ones.
Therefore, the novelty score can be computed from this prediction error — higher errors indicate novel features in unexplored regions, while lower errors suggest familiar features. }
Specifically, the estimator network can be defined as $\psi(\cdot): \mathcal{T}\rightarrow \mathbb{R}^{|\mathcal{T}|}$ and its orthogonal random network is denoted by $\psi^{\perp}(\cdot): \mathcal{T}\rightarrow \mathbb{R}^{|\mathcal{T}|}$, where $\mathcal{T}$ denotes the set of collected sequence and $|\mathcal{T}|$ is the size of $\mathcal{T}$. $\psi(\cdot)$ and $\psi^{\perp}(\cdot)$ following the same architecture as Performance Predictor.



\noindent\textbf{Cold Start:} 
In the cold start stage, the cascading agent will collect sufficient diverse feature transformation sequences. The Performance Predictor will then be optimized by:
\begin{equation}\label{varphi}
    L_{\varphi} = \frac{1}{|\mathcal{T}|} \sum_{T_i\in\mathcal{T}} (\mathcal{A}(T_i(\mathcal{F})) - \varphi(T_i))^2.
\end{equation}
{\color{black}For the Novelty Estimator, we train the estimator network supervised by the output of the target network, given as: }
\begin{equation}\label{psi}
    L_{\psi} = (\psi(\mathcal{T}) - \psi^{\perp}(\mathcal{T}))^2.
\end{equation}
We list the details of the cold start stage in Algorithm~\ref{alg:1}. 
After the cold start, the Performance Predictor and Novelty Estimator will undergo continual finetuning to adapt to evolving data and transformation patterns. The details are introduced in Algorithm~\ref{alg:2}. 
{\color{black} The overall pipeline of cold start and fine-tuning is depicted in Figure~\ref{fintune}.}

\begin{algorithm}[!h]
\SetAlgoLined
\KwInput{dataset $D<\mathcal{F},y>$.}
\KwOutput{performance predictor $\varphi$, novelty estimator $\psi$.}
\textbf{Define:} state $s$, action $a$, reward $r$, discount $\gamma$, feature transformation process $T$, cold start step number $M$, evaluation component training epoch $K$.\\
\textbf{Initialization}: cascaded agents $\pi$, value function $V$, experience replay $\mathcal{H}$, evaluation metric $\mathcal{A}$. \\
    \For{i=1 \textbf{to} M}{
        $T_{i} \gets \pi(T_{i-1}(\mathcal{F}))$;{\hfill\color{gray}\textbf{\#} generate features via RL agents} \\
        $v_{i} \gets \mathcal{A}(T_{i}))$;{\hfill\color{gray}\textbf{\#} evaluation via metrics} \\
        $\mathcal{P}_{i} = \delta_{i} \gets r_i+\gamma V(s_{i+1}) - V(s_i)$;  {\hfill\color{gray}\textbf{\#} TD error}\\
        store$<s_{i},a_{i},r_{i},s_{i+1},T_{i},v_{i}>$ in $\mathcal{H}$ with $\mathcal{P}_{i}$\;
        \If{$\mathcal{H}$ is full}{
            sample transition $m_i:<s_{i},a_{i},r_{i},s_{i+1}> \sim \mathcal{B}$\; {\color{gray}\textbf{\#} $\mathcal{B}$ is the distribution of priority.}\\
            optimize $\pi$ by $m_i$ wrt Actor-Critic loss\;
        }
    }
    Initialize $\varphi$ and $\psi$ \;
    \For{k=1 \textbf{to} K}{
        sample record $(T_{k},v_{k})$ from $\mathcal{H}$ uniformly\;
        train $\varphi$ wrt MSE loss\;
        train $\psi$ wrt MSE loss\;
    }
\caption{Cold Start and Evaluation Components Training}
\label{alg:1}
\end{algorithm}

\begin{algorithm}[!h]
\SetAlgoLined
\KwInput{dataset $D<\mathcal{F},y>$, performance threshold $\alpha$, novelty threshold $\beta$, performance predictor $\varphi$, novelty estimator $\psi$, retrain episode $E$, evaluation component retraining epoch $K$.}
\KwOutput{optimal dataset $D'$.}
\textbf{Define:} state $s$, action $a$, reward $r$, discount $\gamma$, feature
transformation process $T(D)$, value function $V$, current best performance $v'$, exploration step $N$.\\

\For{j=1 \textbf{to} N}{
    $T_{j} \gets \pi(T_{j-1}(\mathcal{F}))$;{\color{gray}\textbf{\#}generate features via RL agents}\\

    ${n}_{j} \gets \psi(T_{j})$;  {\hfill\color{gray}\textbf{\#} estimate new feature set novelty}\\
    $\hat{v}_{j} \gets \varphi(T_{j})$;  {\hfill\color{gray}\textbf{\#} predict new feature set performance}\\
    \If{$n_{j} < \alpha\  and\ \hat{v}_{j} < \beta$}{
    $v_{j} = \hat{v}_{j}$;  {\hfill\color{gray}\textbf{\#} predicted performance as feedback}\\ 
    }
    \Else{
    $v_{j} = \mathcal{A}(T_{j})$; {\hfill\color{gray}\textbf{\#} ML performance as feedback}\\
    }
    \If{$v_{j} > v'$}{
        $v' \gets v_{j},\ D' \gets D_{j}$\;
        }
    $\mathcal{P}_{j} = \delta_{j} \gets r_j+\gamma V(s_{j+1})-V(s_{j})$;  {\color{gray} \textbf{\#} TD error as priority}\\
store$<s_{j},a_{j},r_{j},s_{j+1},a_{j+1},T_{j},v_{j}>$ in $\mathcal{H}$ with $\mathcal{P}_{j}$;{\color{gray}\# update $\mathcal{H}$ based on the sample count and priority $\mathcal{P}$\;}
    sample transition from  priority distribution $m_j \sim \mathcal{B}$\;
    optimize $\pi$ by $m_j$ wrt Actor-Critic loss\;
    \If{$j \equiv 0\ mod\ E$}{
        \For{k=0 \textbf{to} K}{
        sample record $(T_{k},v_{k})$ from $\mathcal{H}$ uniformly\;
        finetune $\varphi$ wrt performance loss\;
        finetune $\psi$ w.r.t. distillation loss\;
        }
    }
}
\caption{Effective Exploration and Continual Training}
\label{alg:2}
\end{algorithm}

\subsection{Efficient Exploration and Optimizaiton} \label{opt}

\noindent\textbf{Reward Estimation.}
We first introduce the critical role of evaluation components within \model\ and how to adaptively integrate them into the training pipeline for an effective reward estimation. 

\noindent\textit{\uline{Reward from Downstream Task}}:
In the cold start stage, the reward for the cascading system will be generated from the evaluation of the downstream tasks.  
As the objective in Equation~\ref{objective}, the reward $r_i$ is calculated as follows:
\begin{equation}
    r_i = \mathcal{A}(T_i(\mathcal{F}),y) - \mathcal{A}(T_{i-1}(\mathcal{F}),y)
\end{equation}
where $T_i$ indicates the transformation sequence at the $i$-th step. 

\noindent\textit{\uline{Pseudo-reward from Evaluation Components.}}
Downstream task evaluation is extremely time-consuming and sparse during exploration. 
Thus, we generate the pseudo-reward from evaluation components for a light optimization target. 
In detail, given the $i$-step transformation sequence $T_i$, the estimated reward will be calculated by:
\begin{equation}\label{reward}
\begin{aligned}
    r_i = \overbracket{(\varphi(T_i) - \varphi(T_{i-1}))}^{\textit{Estimated Performance}} &+ \epsilon_i\overbracket{(\psi(T_i) - \psi^{\perp}(T_i))}^{\textit{Novelty}},\\
    \epsilon_i = \epsilon_e + (\epsilon_s &- \epsilon_e) \times e^{-\frac{i}{M}},
\end{aligned}
\end{equation}
where $\varphi(T_i)$ is the pseudo-performance\footnote{\color{black}Pseudo-performance is the output of the Performance Predictor, which indicates the predicted downstream task performance on the generated dataset.} predicted by the Performance Predictor, and $(\psi(T_i) - \psi^{\perp}(T_i))$ denoted the estimated novelty. 
$\epsilon_i$ represents the weighting factor for the novelty reward in step-$i$, constrained to lie within the interval $\epsilon_i\in [\epsilon_e, \epsilon_s]$. 
$M$ is decay factor, which determines decay rate of $\epsilon_i$. 
The term $\epsilon_i$ guides the agent from the exploration of novel samples (stage with higher weights) and then to the exploration of high-quality novel samples (stage with lower weights).

\noindent\textit{\uline{Adaptively Adopt Two Strategies.}} 
The evaluation components not only act as a supplement to the time-intensive downstream tasks but also can self-monitor and decide when to invoke these downstream evaluations.
If a sequence exhibits high novelty and the Performance Predictor also estimates high downstream task performance—specifically, when the predictions are in the top 
$\alpha$ percentile for performance (potential high performance) or top 
$\beta$ percentile for novelty (unseen sequence)—the downstream task evaluation is triggered. 
This adaptive strategy ensures that the reinforcement learning system focuses its resources efficiently, evaluating downstream tasks only when the potential for significant learning or critical validation exists. 
By leveraging the strengths of both the Performance Predictor and the Novelty Estimator, the system dynamically adjusts its evaluation methodology, optimizing both computational resources and learning efficacy.

\noindent\textbf{Replay Critical Transformation Memory for Optimization.} 

\noindent\textit{\uline{Memory Collection:}} During the cold start stage and efficient exploration stage, the cascading learning system will generate a transformation and collect its correlated memory: $m_i = <s_i, a_i, r_i, s_{i+1}, a_{i+1}, T_i, v_i>$, where $m_i$ denotes the memory of $i$-th exploration step. $s_i$ and $s_{i+1}$ are the state representations (e.g., the input of each agent's policy network) of the current exploration step and the next step, i.e., after transformation. $v_i$ is the estimated or downstream task evaluated performance.

\noindent\textit{\uline{Optimization of Agents:}} The optimization goal for each agent is determined by the expected cumulative reward:
\begin{equation}
    \max_{\pi} \mathbb{E}_{m_i \sim \mathcal{B}} \left[ \sum_{i=0}^{M} \gamma_i r_i \right],
\end{equation}
with $\mathcal{B}$ representing the distribution of memories in the prioritized replay buffer, $\gamma$ being the discount factor, and $M$ representing the temporal horizon.
Furthermore, we define the Q-function, represented by $Q(s, a)$, as the anticipated return from performing action $a$ in state $s$ and subsequently adhering to policy $\pi$:
\begin{equation}
    Q(s, a) = \mathbb{E} \left[ r + \gamma \max_{a'} Q(s', a') \mid s, a \right].
\end{equation}
The loss function in the actor and critic networks during training is performed as follows:
\begin{equation}
\begin{aligned}
    \text{Critic Update:} & \quad L_V = \mathbb{E}_{m_i \sim \mathcal{B}} \left[ \left( V(s_i) - \left( r_i + \gamma V(s_{i+1}) \right) \right)^2 \right], \\
    \text{Actor Update:} & \quad L_\pi = \mathbb{E}_{m_i \sim \mathcal{B}} \left[ \log \pi(a_i | s_i) A(s_i, a_i) \right].
\end{aligned}    
\end{equation}
where $A(s, a) = Q(s, a) - V(s)$ represents the advantage function, facilitating the gradient estimation for policy improvement.

\noindent\textit{\uline{Replay Critical Memories:}}
In \model, learning efficiency can be enhanced by focusing on critical experiences, which can be identified by their high temporal difference (TD) errors.
By giving the memory unit $m_i$ in step-$i$, its priority $\mathcal{P}_i$ and the probability $\mathcal{B}_i$ to select this memory can be derived by: 
\begin{equation}
\begin{aligned}
        \mathcal{P}_i = r_i + &\gamma V(s_{i+1}) - V(s_i),\quad
        \mathcal{B}_i &= \frac{\mathcal{P}_i}{\sum_{k} \mathcal{P}_k}.
\end{aligned}
\end{equation}
With the evolving reward from Equation~\ref{reward}, the framework will first focus on high-novelty samples and then on high-quality samples. 
The finetuning of evaluation components will also extract training samples from this distribution.


\section {Time Complexity Analysis} 
The time complexity of \textit{iterative-feedback} reinforcement learning-based methods mainly includes \textit{agent decision time}, \textit{downstream task feedback time}, and \textit{optimization time}. 
The decision inference and optimization times vary with different reinforcement learning framework selections and are minor overall.
Thus, we focus on optimizing the downstream task feedback process. 
This significant time consumption is associated with multiple training rounds in downstream machine-learning tasks. 
In addition, this data set size will grow exponentially due to the new feature generation.
However, the time complexity of ML tasks usually depends on the number of features and samples.
We replace this process with a Performance Predictor that requires just one forward pass.
The predictor's time complexity is solely based on the number of features, thereby \textbf{reducing time complexity by decoupling the dependency on the total generated dataset}.
Refer to Section~\ref{time.experiment} and~\ref{scalability1} for detailed information on quantitative time analysis.
\section{Experimental Setup}\label{setup}

\begin{table*}[!h]
\centering
\caption{Overall Performance. The best results are highlighted in \textbf{bold}. The second-best results are highlighted in \underline{underline}. We reported F1-Score for classification tasks (C), 1-RAE for regression tasks (R), and AUC for detection tasks (D). \textcolor{black}{The t-statistics and p-values comparing the performances of each baseline with \model are presented in the last two rows.}}
\label{table_overall_perf}
\setlength{\tabcolsep}{1mm}{\resizebox{\linewidth}{!}{
\begin{tabular}{cccccccccccccccccc}
\toprule
Name & Source   & Task & Samples & Features & RFG  & {ERG} & LDA & AFT$^\dagger$   & NFS   & TTG  & DIFER$^\dagger$ & \textcolor{black}{OpenFE} & \textcolor{black}{CAAFE} & GRFG$^\dagger$ & \model$^*$ \\  \midrule
Alzheimers  & Kaggle & C & 2149 & 33 & 0.936 & \underline{0.956} & 0.584 & 0.907 & 0.914 & 0.925 & 0.952 & \textcolor{black}{0.951} & \textcolor{black}{0.945} & 0.953 & $\textbf{0.974}^{\textcolor{black}{\pm0.010}}$\\  
Cardiovascular  & Kaggle & C & 5000 & 12 & \underline{0.720} & 0.709 & 0.561 & 0.712 & 0.710 & 0.709 & 0.712 & \textcolor{black}{0.706} & \textcolor{black}{0.711} & \textbf{0.722} & $\textbf{0.722}^{\textcolor{black}{\pm0.015}}$ \\  
Fetal Health  & Kaggle & C & 2126 & 22 & 0.913 & 0.917 & 0.744 & 0.918 & 0.914 & 0.709 & 0.944 & \textcolor{black}{0.943} & \textcolor{black}{0.945} & \underline{0.951} & $\textbf{0.954}^{\textcolor{black}{\pm0.008}}$\\  
Pima Indian  & UCIrvine & C & 768 & 8 & 0.693 & 0.703 & 0.676 & 0.736 & 0.762 & 0.747 & 0.746 & \textcolor{black}{0.744}  & \textcolor{black}{0.755} & \underline{0.776} & $\textbf{0.789}^{\textcolor{black}{\pm0.035}}$\\  
SVMGuide3  & LibSVM & C & 1243 & 21 & 0.703 & 0.747 & 0.683 & 0.829 & 0.831 & 0.766 & 0.773 & \textcolor{black}{0.831} & \textcolor{black}{0.828} & \underline{0.850} & $\textbf{0.863}^{\textcolor{black}{\pm0.028}}$\\  
Amazon Employee    & Kaggle   & C   & 32769   & 9 & 0.744 & 0.740 & 0.920  & 0.943 & 0.935 & 0.806 & 0.937 & \textcolor{black}{0.944} & \textcolor{black}{0.943} & \underline{0.946} & $\textbf{0.951}^{\textcolor{black}{\pm0.002}}$ \\  
German Credit      & UCIrvine & C   & 1001    & 24  & 0.695 & 0.661 & 0.627  & 0.751 & \underline{0.765} & 0.731 & 0.752 & \textcolor{black}{0.757} & \textcolor{black}{0.759} & 0.763 & $\textbf{0.777}^{\textcolor{black}{\pm0.015}}$ \\  
Wine Quality Red   & UCIrvine & C   & 999     & 12  & 0.599 & 0.611 & 0.600  & 0.658 & 0.666 & 0.647 & 0.675 & \textcolor{black}{0.658} & \textcolor{black}{0.677} & \underline{0.686} & $\textbf{0.692}^{\textcolor{black}{\pm0.031}}$\\
Wine Quality White & UCIrvine & C   & 4898    & 12    & 0.552 & 0.587 & 0.571  & 0.673 & 0.679 & 0.638 & 0.681 & \textcolor{black}{0.670}  & \textcolor{black}{0.676} & \underline{0.685} & $\textbf{0.691}^{\textcolor{black}{\pm0.005}}$ \\
Jannis & AutoML & C & 83733 & 55 & \underline{0.714} & 0.712 & 0.477 & 0.695 & \underline{0.714} & 0.711 & $\times$ & 0.708 & 0.698 & $\times$ & $\textbf{0.722}^{\pm0.003}$\\
\textcolor{black}{Adult} & \textcolor{black}{AutoML} & \textcolor{black}{C} & \textcolor{black}{34190} & \textcolor{black}{25} & \textcolor{black}{0.837} & \textcolor{black}{0.842} & \textcolor{black}{0.804} & \textcolor{black}{0.841} & \textcolor{black}{0.827} & \textcolor{black}{0.829} & \textcolor{black}{$\times$} & \textcolor{black}{0.832} & \textcolor{black}{\underline{0.849}} & \textcolor{black}{$\times$} & $\textcolor{black}{\textbf{0.850}^{\pm0.051}}$\\
\textcolor{black}{Volkert} & \textcolor{black}{AutoML} & \textcolor{black}{C} & \textcolor{black}{58310} & \textcolor{black}{181} & \textcolor{black}{0.623} & \textcolor{black}{0.609} & \textcolor{black}{0.418} & \textcolor{black}{0.598} & \textcolor{black}{0.602} & \textcolor{black}{0.656} & \textcolor{black}{$\times$} & \textcolor{black}{\underline{0.671}} & \textcolor{black}{0.652} & \textcolor{black}{$\times$} & $\textcolor{black}{\textbf{0.672}^{\pm0.012}}$\\
\textcolor{black}{Albert} & \textcolor{black}{AutoML} & \textcolor{black}{C} & \textcolor{black}{425240} & \textcolor{black}{79} & \textcolor{black}{\underline{0.678}} & \textcolor{black}{0.619} & \textcolor{black}{0.530} & \textcolor{black}{$\times$} & \textcolor{black}{0.662} & \textcolor{black}{0.667} & \textcolor{black}{$\times$} & \textcolor{black}{0.676} & \textcolor{black}{0.668} & \textcolor{black}{$\times$} & $\textcolor{black}{\textbf{0.686}^{\pm0.004}}$\\
\midrule
OpenML\_618        & OpenML   & R   & 1000    & 50  & 0.415 & 0.427  & 0.372 & 0.665 & 0.640 & 0.587 & 0.644 & \textcolor{black}{0.717} & \underline{0.725} & 0.672 & $\textbf{0.786}^{\textcolor{black}{\pm0.015}}$ \\  
OpenML\_589        & OpenML   & R   & 1000    & 25  & 0.638 & 0.560 & 0.331  & 0.672 & 0.711 & 0.682 & 0.715 & \textcolor{black}{0.719} & \textcolor{black}{0.714} & \underline{0.753} & $\textbf{0.768}^{\textcolor{black}{\pm0.017}}$ \\ 
OpenML\_616        & OpenML   & R   & 500    & 50    & 0.448 & 0.372 & 0.385  & 0.585 & 0.593 & 0.559 & 0.556 & \textcolor{black}{0.632} & \underline{0.647} & 0.603 & $\textbf{0.726}^{\textcolor{black}{\pm0.031}}$\\  
OpenML\_607        & OpenML   & R   & 1000    & 50    & 0.579 & 0.406 & 0.376  & 0.658 & 0.675 & 0.639 & 0.636 & \underline{0.730} & \textcolor{black}{0.651} & 0.680 & $\textbf{0.764}^{\textcolor{black}{\pm0.043}}$\\  
OpenML\_620        & OpenML   & R   & 1000    & 25   & 0.575 & 0.584 & 0.425 & 0.663 & 0.698 & 0.656 & 0.639 & \textcolor{black}{0.689} & \textcolor{black}{0.701} & \underline{0.714} & $\textbf{0.737}^{\textcolor{black}{\pm0.007}}$\\  
OpenML\_637        & OpenML   & R   & 500    & 50    & 0.561 & 0.497 & 0.494  & 0.564 & 0.581 & 0.575 & 0.549 & \underline{0.591} & \textcolor{black}{0.576} & 0.589 & $\textbf{0.700}^{\textcolor{black}{\pm0.043}}$\\  
OpenML\_586       & OpenML   & R   & 1000    & 25   & 0.595 & 0.546 & 0.472  & 0.687 & 0.748 & 0.704 & 0.665 & \textcolor{black}{0.745} & \textcolor{black}{0.687} & \underline{0.783} & $\textbf{0.807}^{\textcolor{black}{\pm0.020}}$\\  
\midrule
WBC        & UCIrvine   & D   & 278    & 30    & 0.753 & 0.766 & 0.736  & 0.743 & 0.755 & 0.752 & \underline{0.956} & \textcolor{black}{0.905
} & \textcolor{black}{0.601} & 0.785 & $\textbf{0.972}^{\textcolor{black}{\pm0.058}}$\\
Mammography        & OpenML   & D   & 11183    & 6     & 0.731 & 0.728 & 0.668  & 0.714 & 0.728 & 0.734 & 0.532 & \textcolor{black}{\underline{0.806}} & \textcolor{black}{0.668} & 0.751 & $\textbf{0.860}^{\textcolor{black}{\pm0.036}}$\\   
Thyroid        & UCIrvine   & D   & 3772    & 6    & 0.813 & 0.790 & 0.778  & 0.797 & 0.722 & 0.720 & 0.613 & \underline{0.967} & \textcolor{black}{0.776} & 0.954 & $\textbf{0.999}^{\textcolor{black}{\pm0.008}}$ \\ 
SMTP        & UCIrvine   & D   & 95156    & 3      & 0.885 & 0.836 & 0.765  & 0.881 & 0.816 & 0.895 & 0.573 & \textcolor{black}{0.494} & \textcolor{black}{0.732} & \underline{0.943} & $\textbf{0.950}^{\textcolor{black}{\pm0.061}}$ \\
\arrayrulecolor{black}
  \bottomrule
  \toprule
\textcolor{black}{T-stat} & \textcolor{black}{$-$} & \textcolor{black}{$-$} & \textcolor{black}{$-$} & \textcolor{black}{$-$} & \textcolor{black}{6.567} & \textcolor{black}{6.491} & \textcolor{black}{9.679} & \textcolor{black}{6.159} & \textcolor{black}{5.115} & \textcolor{black}{6.214} & \textcolor{black}{4.093} & \textcolor{black}{3.003} & \textcolor{black}{3.992} & \textcolor{black}{3.793} & \textcolor{black}{$-$}\\
\textcolor{black}{P-value} & \textcolor{black}{$-$} & \textcolor{black}{$-$} & \textcolor{black}{$-$} & \textcolor{black}{$-$} & \textcolor{black}{5.310e-7} & \textcolor{black}{6.345e-7} & \textcolor{black}{7.051e-10} & \textcolor{black}{1.685e-6} & \textcolor{black}{1.754e-5} & \textcolor{black}{1.218e-6} & \textcolor{black}{3.100e-4} & \textcolor{black}{3.175e-3} & \textcolor{black}{3.075e-4} & \textcolor{black}{6.139e-4} & \textcolor{black}{$-$}
\\
  \bottomrule
\arrayrulecolor{black}
\end{tabular}}}
\begin{tablenotes}
    \item \textcolor{black}{* The standard deviation is computed based on the results of 5 independent runs.
    \item $\dagger$ Methods marked by "$\times$" indicate that their execution time is unacceptably prolonged on some of the selected datasets.}
\end{tablenotes}
\vspace{-0.5cm}
\end{table*}
\noindent\textbf{Dataset Description.}
\textcolor{black}{We adopt 23 publicly available datasets covering various domains, including real-world scenarios, medical care, life sciences, and anomaly detection from Kaggle~\cite{kaggle}, UCIrvine~\cite{uci}, LibSVM~\cite{libsvm}, OpenML~\cite{openml} and AutoML~\cite{guyon10analysis}.These datasets comprise 12 classification tasks (C), 7 regression tasks (R) and 4 detection tasks (D). More details of these datasets are listed in Table~\ref{table_overall_perf}.}

\noindent\textbf{Evaluation Metrics.}
To comprehensively evaluate the performance of our method, considering the differences between various downstream tasks, we utilize widely adopted evaluation metrics~\cite{kdd2022,xiao2024traceable} for each task type. 
For classification tasks, we use F1-score, Precision, and Recall as evaluation metrics. 
For regression tasks, we use 1 - Relative Absolute Error (1-RAE), 1 - Mean Absolute Error (1-MAE) and 1 - Mean Squared Error (1-MSE). 
For detection tasks, we use Precision, F1-score, and Area Under ROC Curve (AUC) as evaluation metrics. 

\noindent\textbf{Baseline Methods.}
We compare our method with 10 widely-used feature transformation methods: 
(1) \textbf{RFG} generates new features by randomly selecting candidate features and operations. 
(2) \textbf{ERG} applies operations to all features to expand the feature space, then selects key features. 
(3) \textbf{LDA}~\cite{blei2003latent} is a dimensionality reduction technique that projects data into a lower-dimensional space.
(4) \textbf{AFT}~\cite{horn2019autofeat} generates and selects features iteratively to enhance downstream task performance by minimizing redundancy and optimizing feature space exploration.
(5) \textbf{NFS}~\cite{chen2019neural} employs a recurrent neural network-based controller, trained with reinforcement learning, to generate and optimize feature transformations. 
(6) \textbf{TTG}~\cite{khurana2018feature} explores a transformation graph using reinforcement learning to systematically enumerate feature transformation options.
(7) \textbf{DIFER}~\cite{zhu2022difer} embeds feature crossover sequences and uses greedy search to identify and optimize transformed features in a continuous vector space.
(8) \textbf{OpenFE}~\cite{zhang2023openfe} automates feature generation by using a novel feature boosting method and a two-stage pruning algorithm, achieving high efficiency and accuracy in identifying effective features for tabular data.
(9) \textbf{CAAFE}~\cite{hollmann2023large} is a context-aware automated feature engineering method for tabular data that leverages large language models to generate semantically meaningful features based on dataset descriptions iteratively.
(10) \textbf{GRFG}~\cite{xiao2024traceable} nested feature generation and selection via cascading reinforcement learning. 

\noindent\textbf{Hyperparameter and Reproducibility.}
All experiments were conducted using five-fold cross-validation, with the training set and test set split in a 4:1 ratio. The reported results are the averages from five independent runs. 
(1) \textit{Reinforcement Learning}.
The reinforcement agents explore for 200 episodes, with each episode consisting of 15 steps.
The Cold Start phase ends at the 10th episode.
Both the Empirical Performance Predictor and the Novelty Estimator re-train every 5 episodes thereafter. 
The thresholds for triggering downstream tasks are set to 10 for \(\alpha\) and 5 for \(\beta\).
For the reward calculation, the novelty reward weight starts at 0.1, decreases in 1000 steps and ends at 0.005.
(2) \textit{Prioritized Experience Replay}.
We used an experience replay memory size of 16.
(3) \textit{Performance Predictor}.
The Performance Predictor consists of 2 stacked LSTM layers with an embedding dimension of 32, followed by 2 fully connected layers with output dimensions of 16 and 1, respectively.
(4) \textit{Novelty Estimator}.
In the Novelty Estimator, the random network and the estimator network utilize the same structural encoder as the Performance Predictor. 
Differently, the random network includes 1 fully connected layer with an output dimension of 1, while the estimator network has 3 fully connected layers with output dimensions of 16, 4, and 1. 
The coupled orthogonal initialization scaling factor is set to 16.0.

\noindent\textbf{Environmental Settings.}
All experiments are conducted on the Ubuntu 18.04.6 LTS operating system, AMD EPYC 7742 64-Core Processor @2.25 GHz, and 8 x NVIDIA A100 GPU with 40G RAM. 
We run all experiments with Python 3.10 and Pytorch~\cite{pytorch} 1.13.1.

\begin{figure*}[!h]
    \centering
    \begin{minipage}{0.24\textwidth}
        \centering
        \includegraphics[width=1\linewidth]{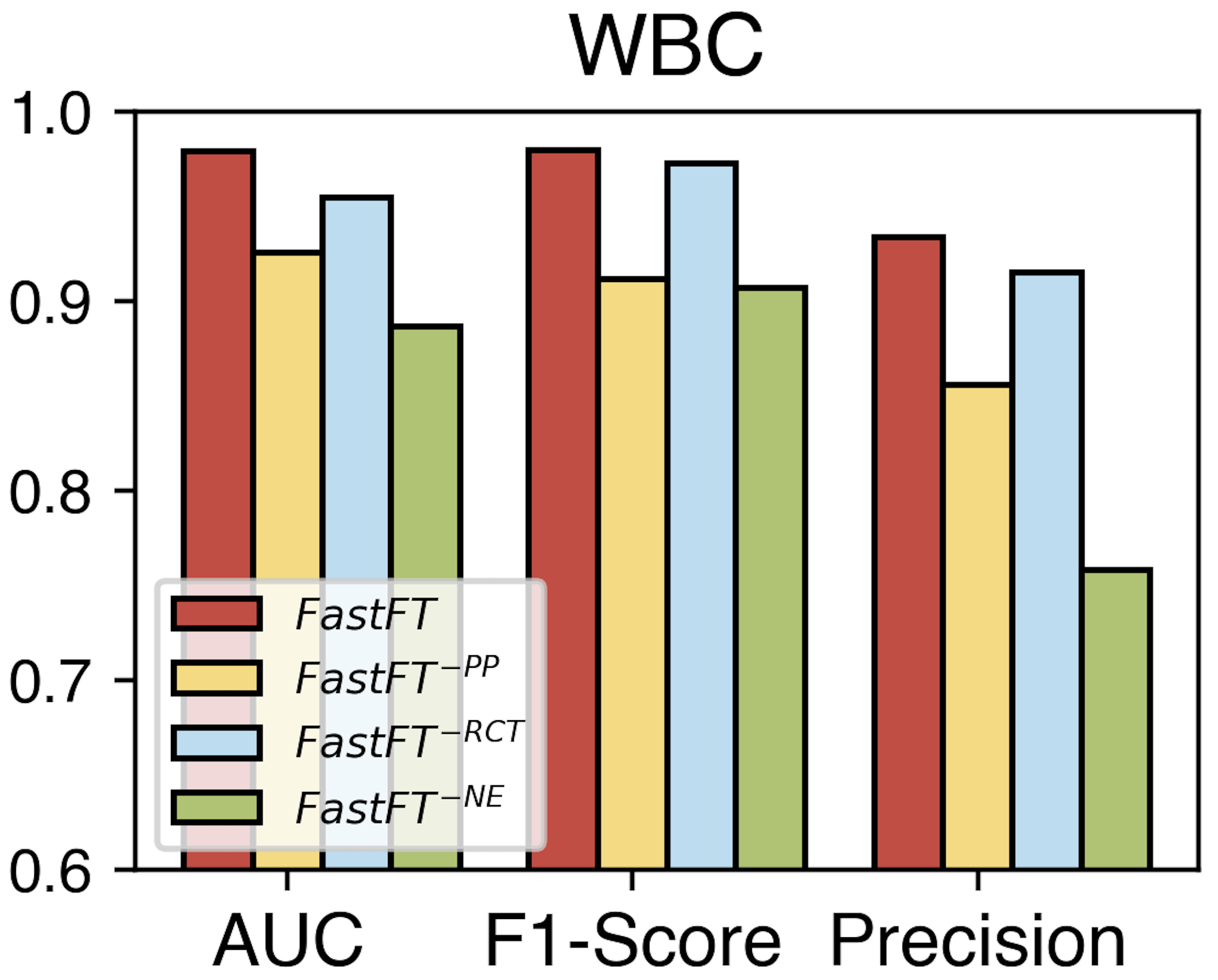}
    \end{minipage}
    \begin{minipage}{0.24\textwidth}
        \centering
        \includegraphics[width=1\linewidth]{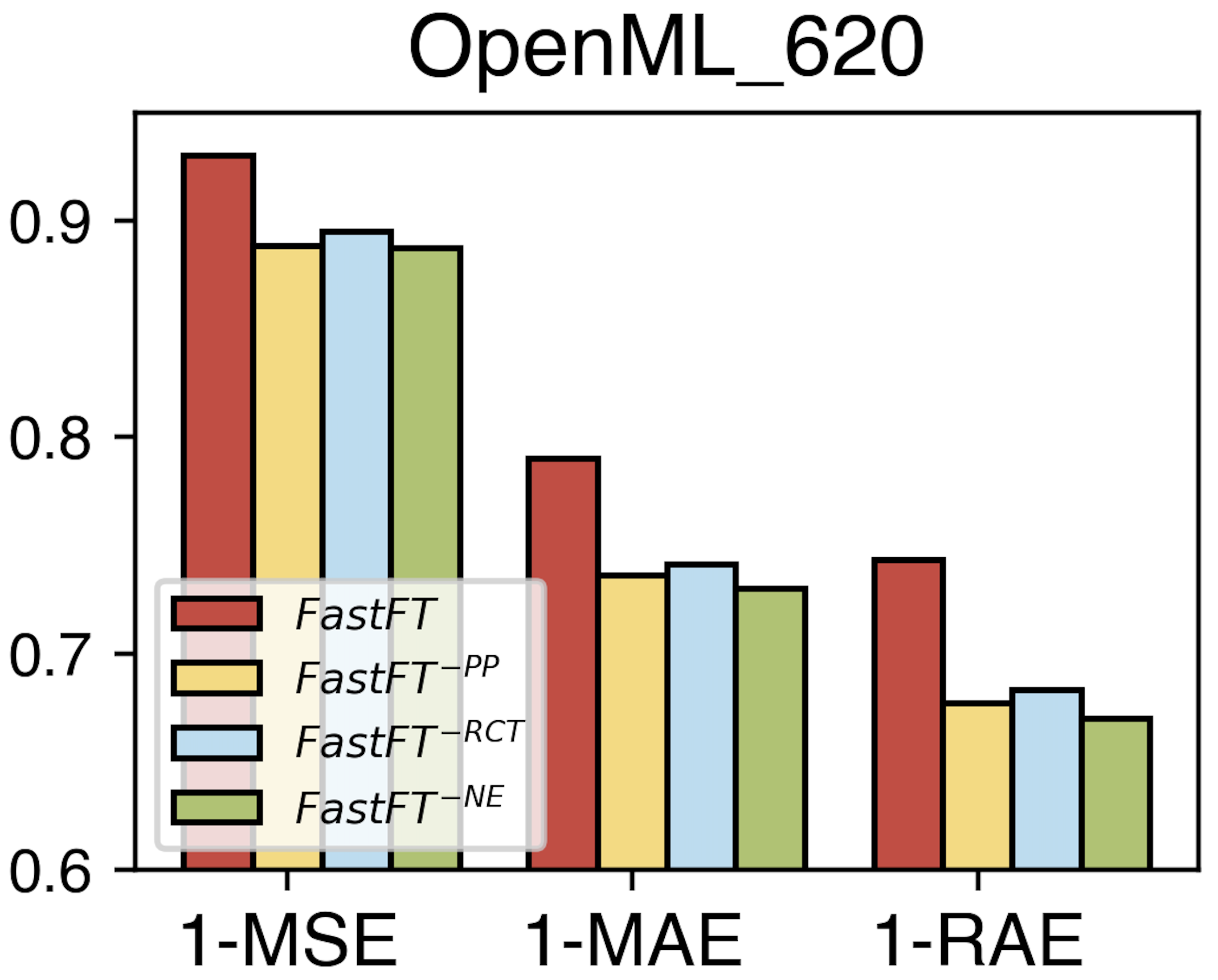}
    \end{minipage}
    \begin{minipage}{0.24\textwidth}
        \centering
        \includegraphics[width=1\linewidth]{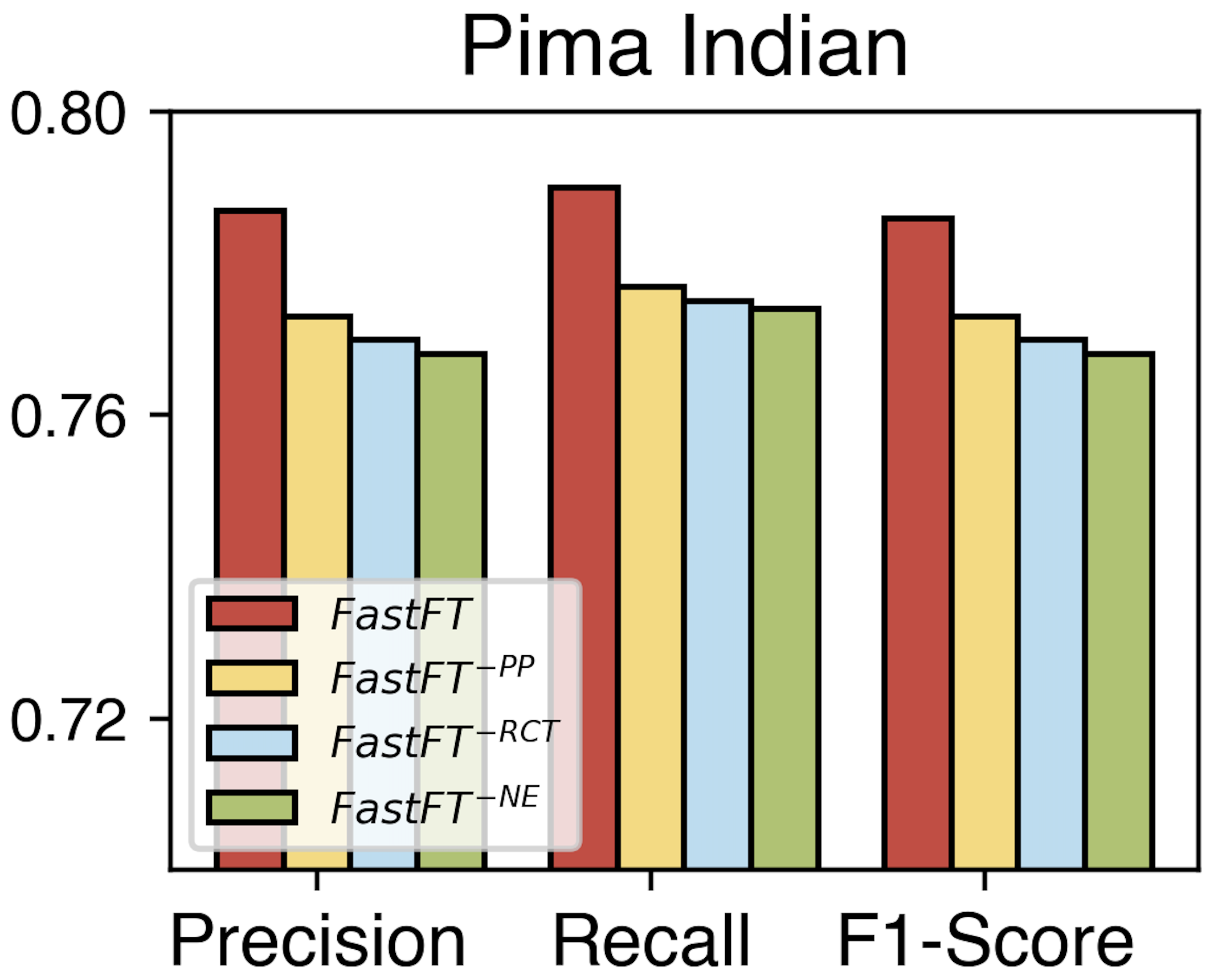}
    \end{minipage}
    \begin{minipage}{0.24\textwidth}
        \centering
        \includegraphics[width=1\linewidth]{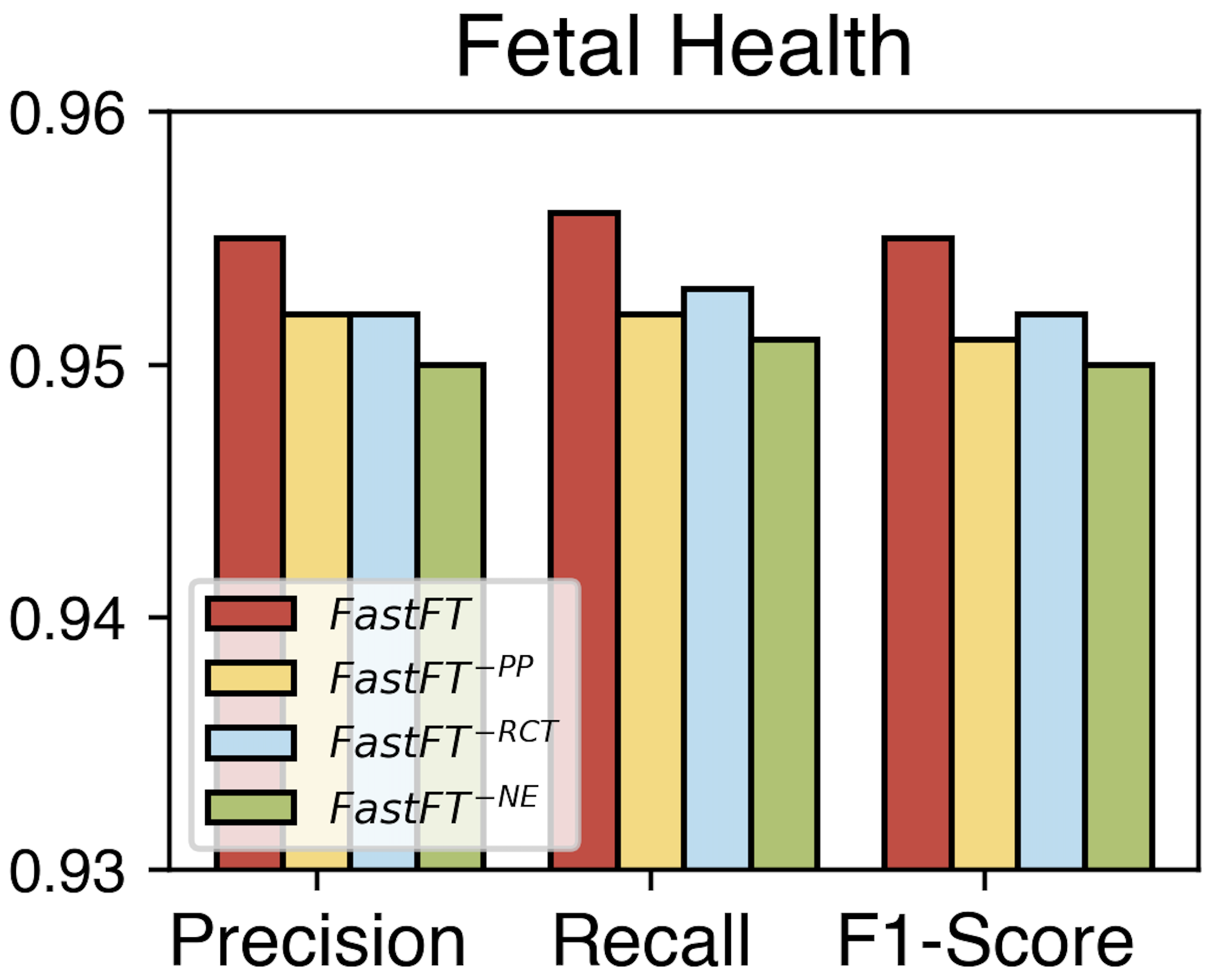}
    \end{minipage}
    \caption{\textcolor{black}{The ablation study of the Performance Predictor, the prioritized experience replay strategy, and the Novelty Estimator}}
    \label{fig:ablation}
\end{figure*}
\begin{table*}[!ht]
\centering
\caption{Time consumption comparison between \model{} and \(\model^{-PP}\) in four datasets with different size. We highlight the consumed time of \model\ in bold and report the average runtime within one episode in terms of seconds. }
\label{tab:time}
\setlength\tabcolsep{16pt} 
\resizebox{\linewidth}{!}{
\begin{tabular}{lcccccccc}
\toprule
{Dataset (Size$^1$)} & \multicolumn{2}{c}{SVMGuide3 (26,103)} & \multicolumn{2}{c}{Wine Quality White (58,800)} & \multicolumn{2}{c}{Cardiovascular (60,000)} & \multicolumn{2}{c}{Amazon Employee (294,921)} \\ \midrule
Method & \cellcolor[HTML]{ECF4FF}\model{}$^{-PP}$& \cellcolor[HTML]{FCE4D6}\model{}& \cellcolor[HTML]{ECF4FF}\model{}$^{-PP}$& \cellcolor[HTML]{FCE4D6}\model{}& \cellcolor[HTML]{ECF4FF}\model{}$^{-PP}$ & \cellcolor[HTML]{FCE4D6}\model{} & \cellcolor[HTML]{ECF4FF}\model{}$^{-PP}$ & \cellcolor[HTML]{FCE4D6}\model{} \\
\midrule
Optimization & \cellcolor[HTML]{ECF4FF}3.31& \cellcolor[HTML]{FCE4D6}\textbf{3.69}& \cellcolor[HTML]{ECF4FF}1.68& \cellcolor[HTML]{FCE4D6}\textbf{2.17}& \cellcolor[HTML]{ECF4FF}2.95 & \cellcolor[HTML]{FCE4D6}\textbf{3.30} & \cellcolor[HTML]{ECF4FF}3.40 & \cellcolor[HTML]{FCE4D6}\textbf{3.76} \\
Estimation & \cellcolor[HTML]{ECF4FF}- & \cellcolor[HTML]{FCE4D6}\textbf{9.64}& \cellcolor[HTML]{ECF4FF}- & \cellcolor[HTML]{FCE4D6}\textbf{4.65} & \cellcolor[HTML]{ECF4FF}- & \cellcolor[HTML]{FCE4D6}\textbf{5.40} & \cellcolor[HTML]{ECF4FF}- & \cellcolor[HTML]{FCE4D6}\textbf{8.76} \\
Evaluation & \cellcolor[HTML]{ECF4FF}51.73& \cellcolor[HTML]{FCE4D6}\textbf{8.20$^{\textbf{\tiny-84.15\%}}$}& \cellcolor[HTML]{ECF4FF}194.24 & \cellcolor[HTML]{FCE4D6}\textbf{34.45$^{\textbf{\tiny-82.26\%}}$}& \cellcolor[HTML]{ECF4FF}114.15 & \cellcolor[HTML]{FCE4D6}\textbf{19.12$^{\textbf{\tiny-83.25\%}}$} & \cellcolor[HTML]{ECF4FF}1111.58 & \cellcolor[HTML]{FCE4D6}\textbf{195.18$^{\textbf{\tiny-82.44\%}}$} \\
Overall & \cellcolor[HTML]{ECF4FF}55.04& \cellcolor[HTML]{FCE4D6}\textbf{21.53$^{\textbf{\tiny-60.88\%}}$}& \cellcolor[HTML]{ECF4FF}195.92& \cellcolor[HTML]{FCE4D6}\textbf{41.27$^{\textbf{\tiny-78.94\%}}$}& \cellcolor[HTML]{ECF4FF}117.1 & \cellcolor[HTML]{FCE4D6}\textbf{27.82$^{\textbf{\tiny-76.24\%}}$}& \cellcolor[HTML]{ECF4FF}1114.98 & \cellcolor[HTML]{FCE4D6}\textbf{207.70$^{\textbf{\tiny-81.37\%}}$} \\ \bottomrule
\end{tabular}         
}
\begin{tablenotes}
\footnotesize
\item[*] $^1$ The data size indicates the number of `\#Sample $\times$ \#Feature'.
\end{tablenotes}
\vspace{-0.6cm}
\end{table*}

\section{Experimental Results}
\subsection{Overall Comparison}\label{overall}
Table~\ref{table_overall_perf} presents the performance of our method on each dataset. 
We observed that \model\ significantly outperforms most random-based or expansion-reduction-based feature transformation methods. 
The primary underlying principle is that the reinforcement learning agent is capable of optimizing the decision-making process, exhibiting greater proficiency compared to random based or expansion-reduction based approaches. 
Another observation is that our method outperforms other iterative-feedback-based methods, such as NFS, TTG, and GRFG. 
A potential reason is that \model\ optimizes the decision-making process not only by the downstream task performance but also by incorporating the novelty of the generated features. 
This enables the cascading agents to explore the feature space more effectively, thereby enhancing the model's performance. 
\textcolor{black}{Additionally, consistently positive t-statistics and p-values well below the 0.05 between our methods and each baseline confirm the statistical superiority of our approach.} 
Overall, this experiment demonstrates the effectiveness of \model\ across multiple datasets and downstream tasks. 

\begin{figure}[!t]
    \centering
    \begin{minipage}{0.24\textwidth}
        \centering
        \includegraphics[width=1\linewidth]{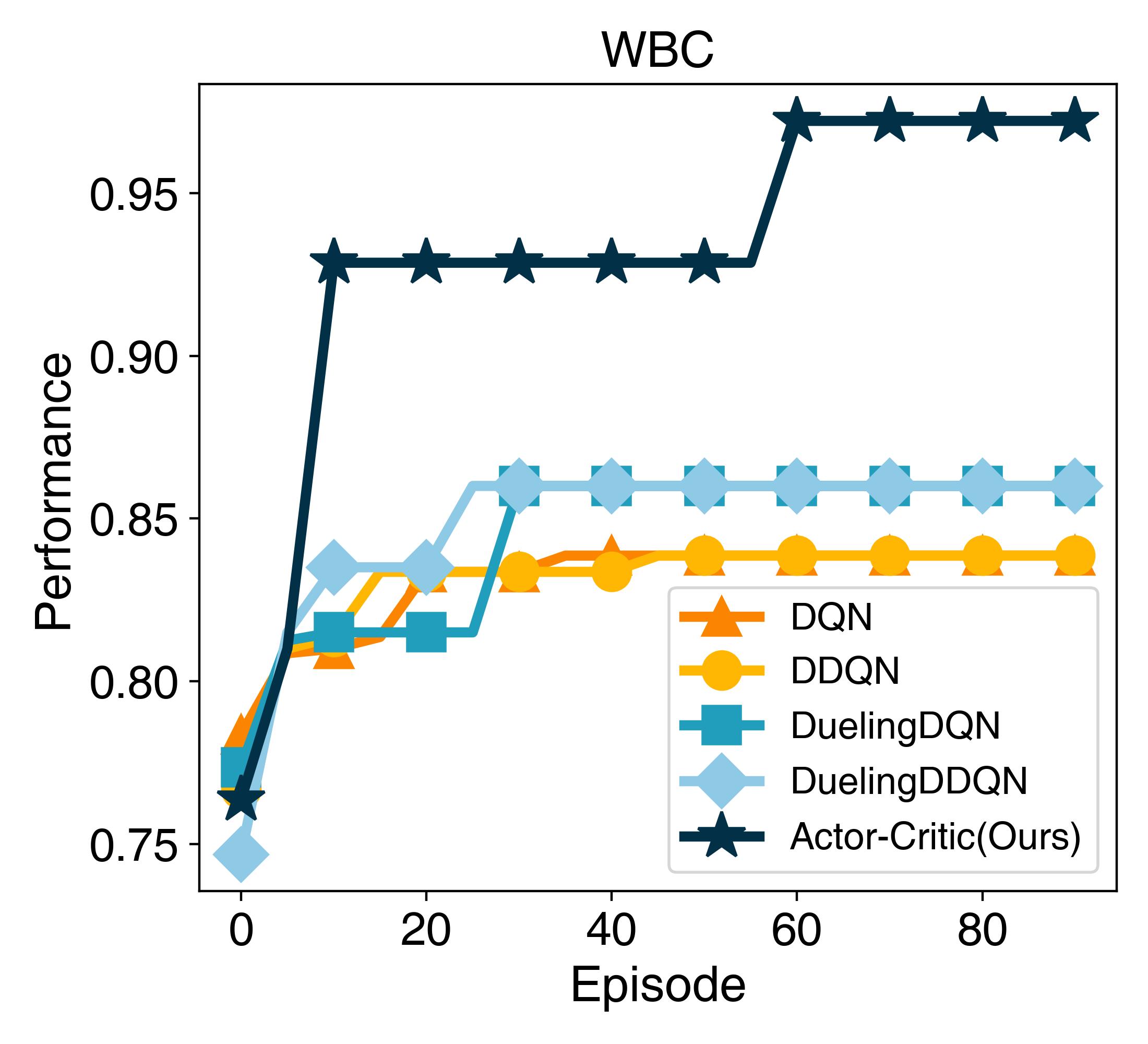}
    \end{minipage}
    \begin{minipage}{0.24\textwidth}
        \centering
        \includegraphics[width=1\linewidth]{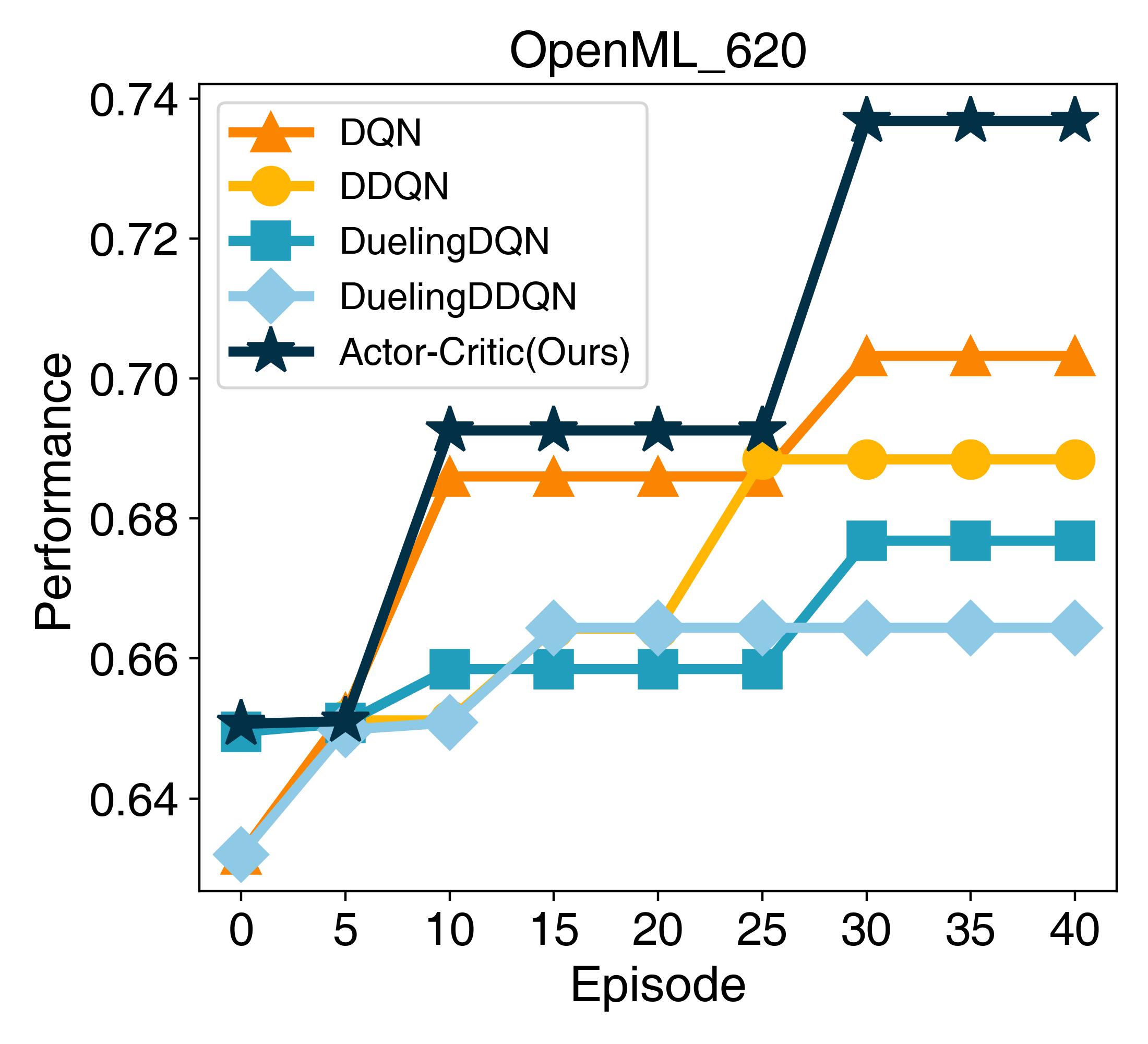}
    \end{minipage}
    \caption{\textcolor{black}{Comparison between reinforcement learning algorithms.}}
    \label{fig:rl_ablation}
    \vspace{-0.5cm}
\end{figure}

\subsection{Study of the Technical Components} \label{ablation}

We conducted the ablation study across four datasets, including three task types and two dataset size.
{\textcolor{black}

\noindent\textbf{Impact of Performance Predictor.}
We designed \({\model}^{-PP}\), which ablates the Performance Predictor component and evaluates all generated feature sets with downstream tasks. From Figure~\ref{fig:ablation}, we can observe a minor performance impact. 
However, Table~\ref{tab:time} reveals that the runtime is significantly reduced when using the Performance Predictor.
We can also observe that the primary runtime savings come from avoiding the expensive evaluation step, which involves running full downstream tasks for the generated dataset. 
Instead, we can usually predict rewards by training the Performance Predictor on only the most critical memory components.
Our approach maintains similar performance levels to evaluating every generated feature set with downstream tasks and enhances computational efficiency.

\noindent\textbf{Impact of Replay Critical Transformation. }
We designed \({\model}^{-RCT}\), which uniformly samples the transformation memories instead of replaying critical ones from the memory buffer. 
From Figure~\ref{fig:ablation}, we observed that critical replay transformation improves model performance. 
This improvement is likely due to replaying critical memory, which allows the agent to focus on the most informative experiences, accelerating the learning process and improving overall performance.

\noindent\textbf{Impact of Novelty Estimator. } 
We designed \({\model}^{-NE}\), which ablates the Novelty Estimator so that the reinforcement learning reward solely depends on the performance score. 
Compared to \({\model}^{-NE}\), we observed that the Novelty Estimator enhances the performance of downstream tasks on the generated feature set. 
The primary reason is that \model\ introduces novelty as the feedback signal, encouraging the model to explore a larger feature transformation space. 
Therefore, it prevents the agent from getting stuck in local optima, improving overall performance.

}
\begin{figure}[!t]
    \centering
    \begin{minipage}{0.24\textwidth}
        \centering
        \includegraphics[width=1\linewidth]{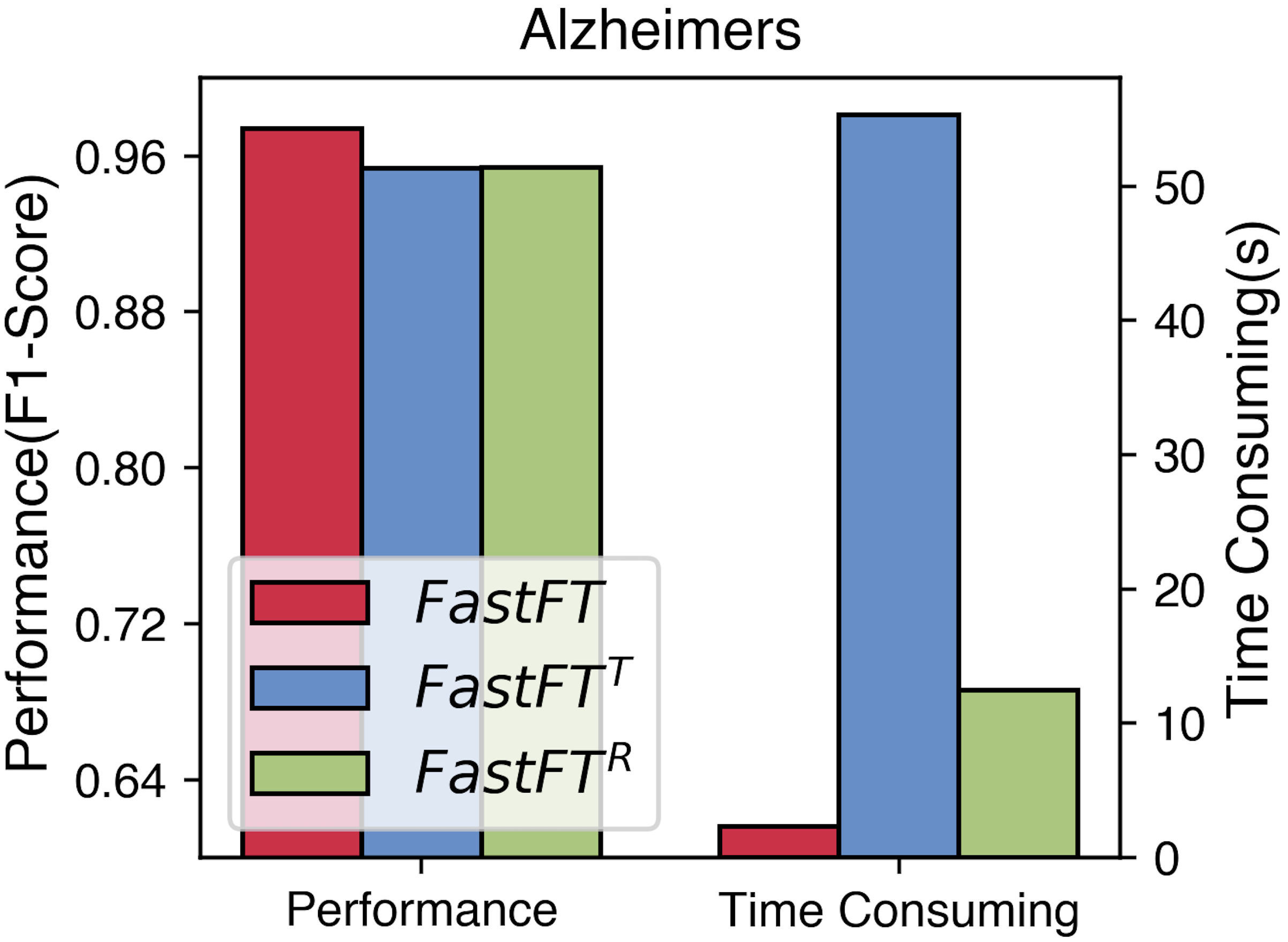}
    \end{minipage}
    \begin{minipage}{0.24\textwidth}
        \centering
        \includegraphics[width=1\linewidth]{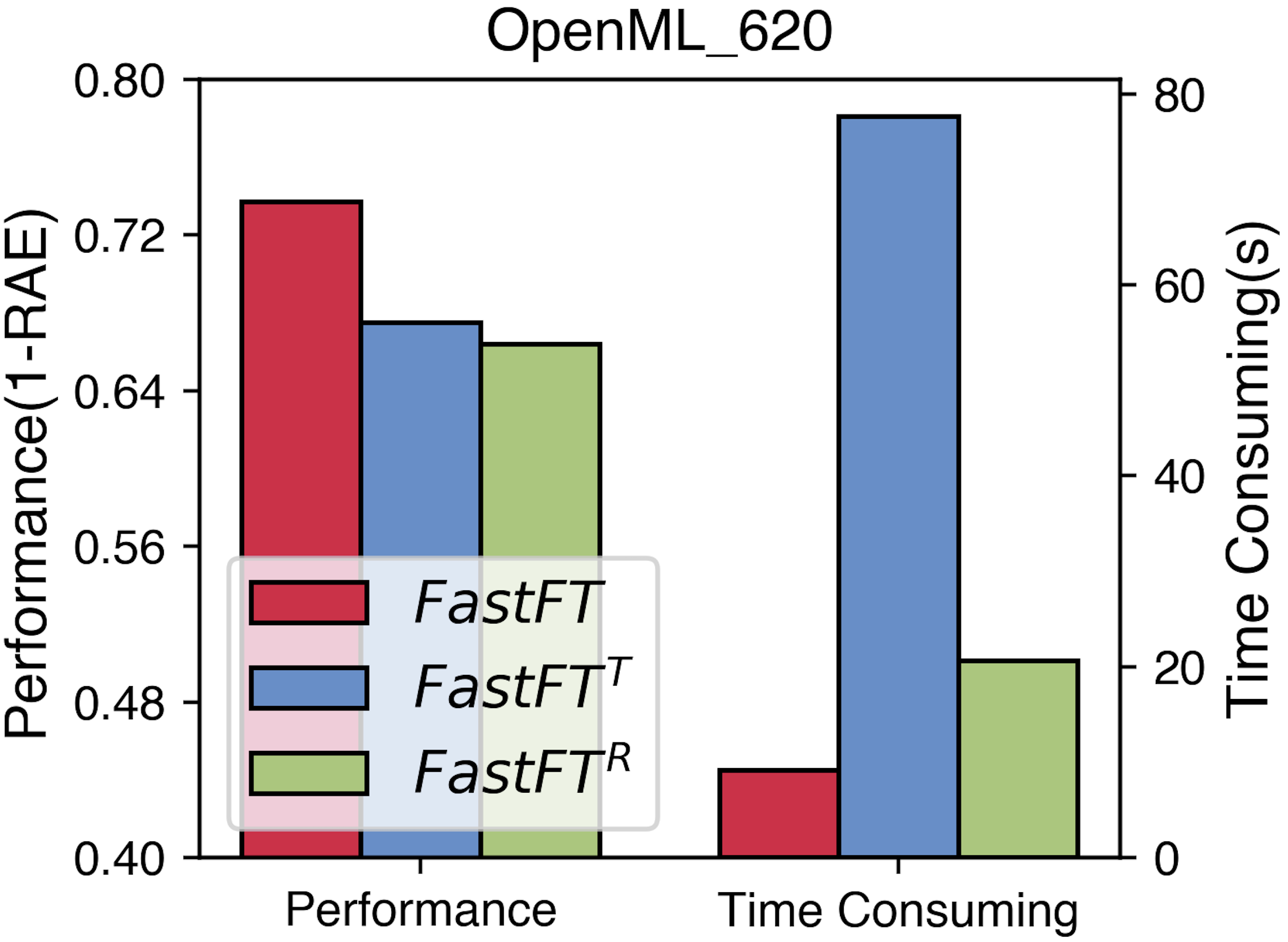}
    \end{minipage}
    \caption{\textcolor{black}{Performance and training time of \model\ using different sequential modeling methods.}}
    \vspace{-0.5cm}
\label{fig:sequence}
\end{figure}
\begin{figure*}[!ht]
    \centering
    \begin{minipage}{0.24\textwidth}
        \centering
        \includegraphics[width=1\linewidth]{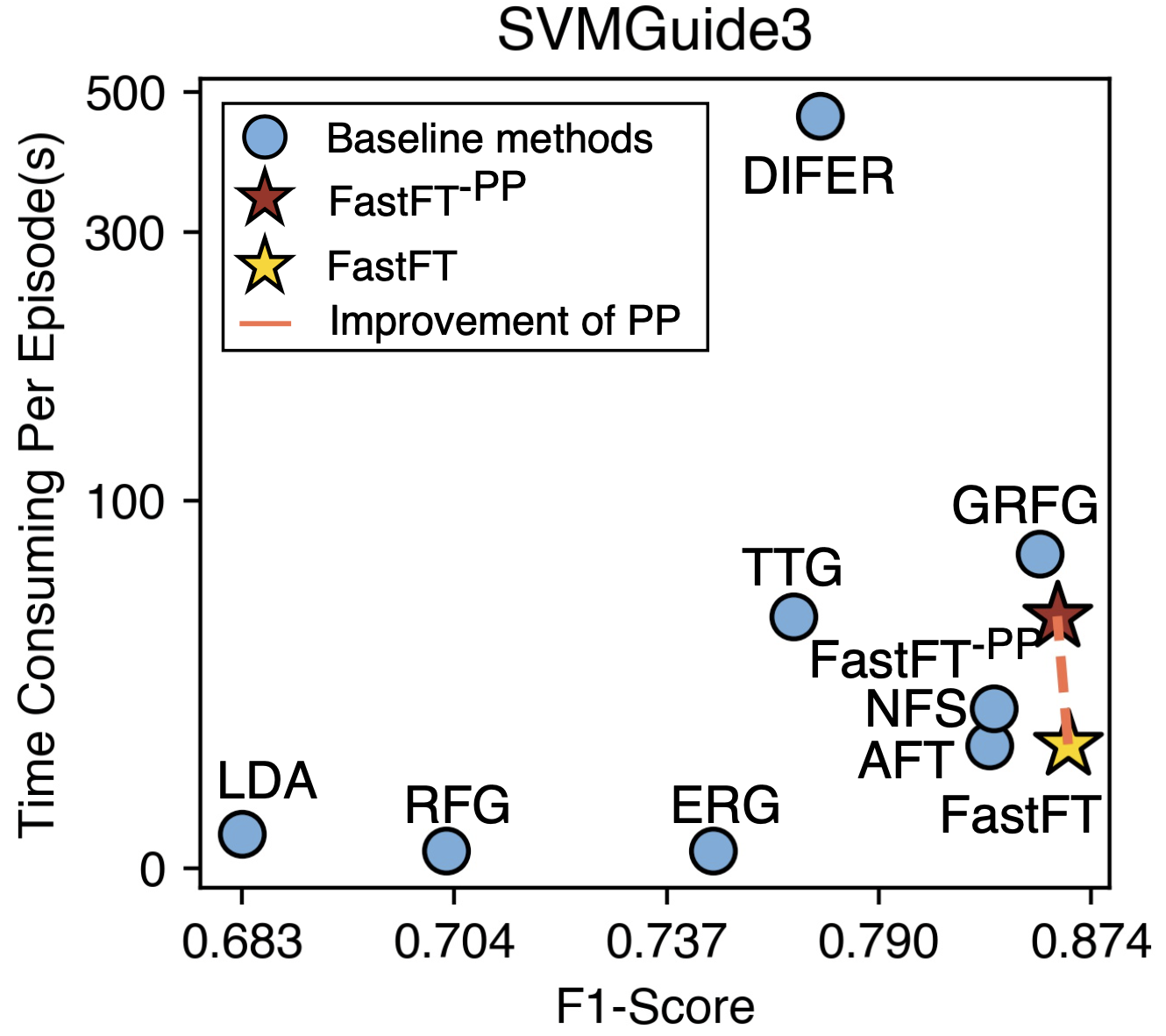}
    \end{minipage}
    \begin{minipage}{0.24\textwidth}
        \centering
        \includegraphics[width=1\linewidth]{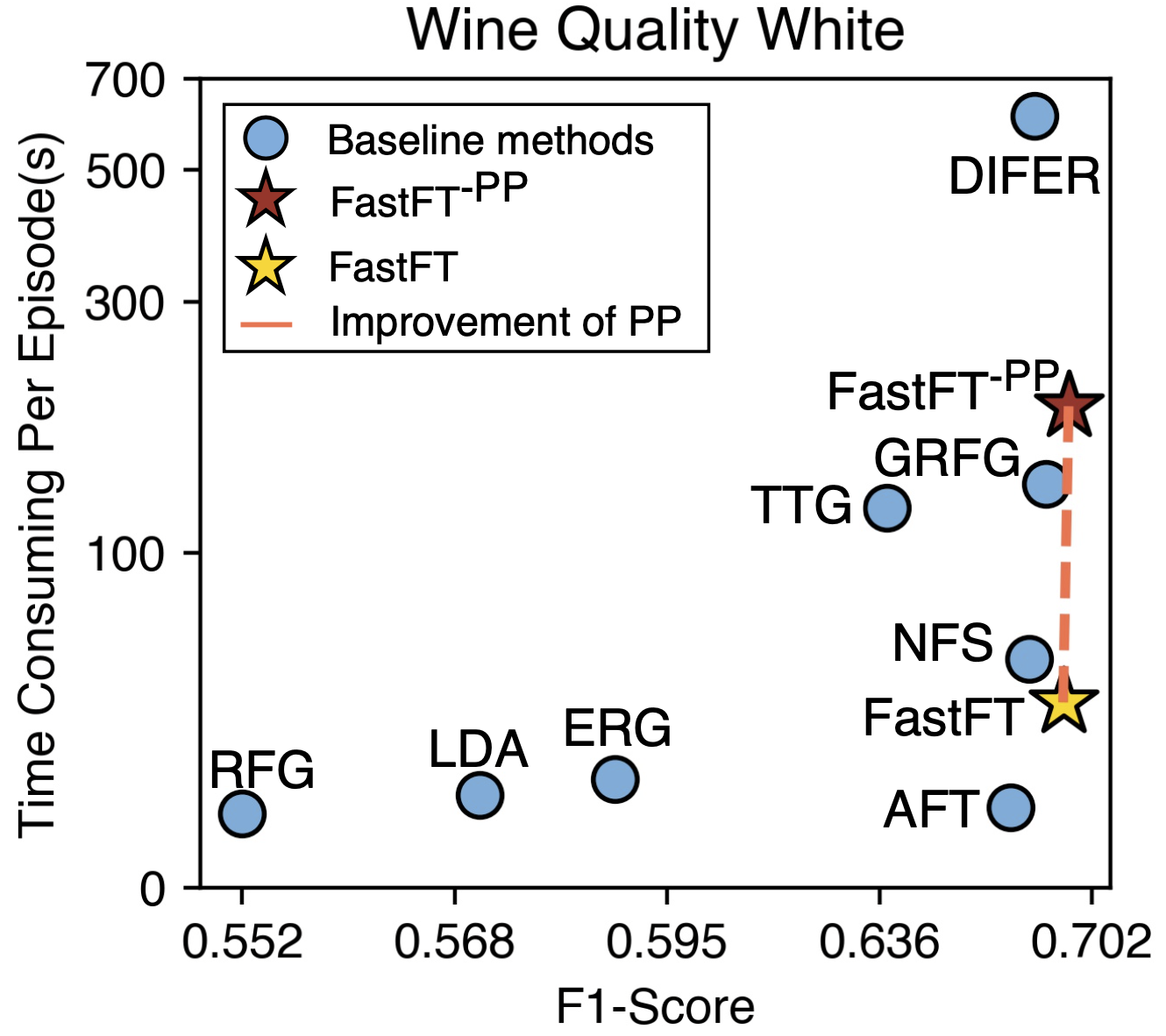}
    \end{minipage}
    \begin{minipage}{0.24\textwidth}
        \centering
        \includegraphics[width=1\linewidth]{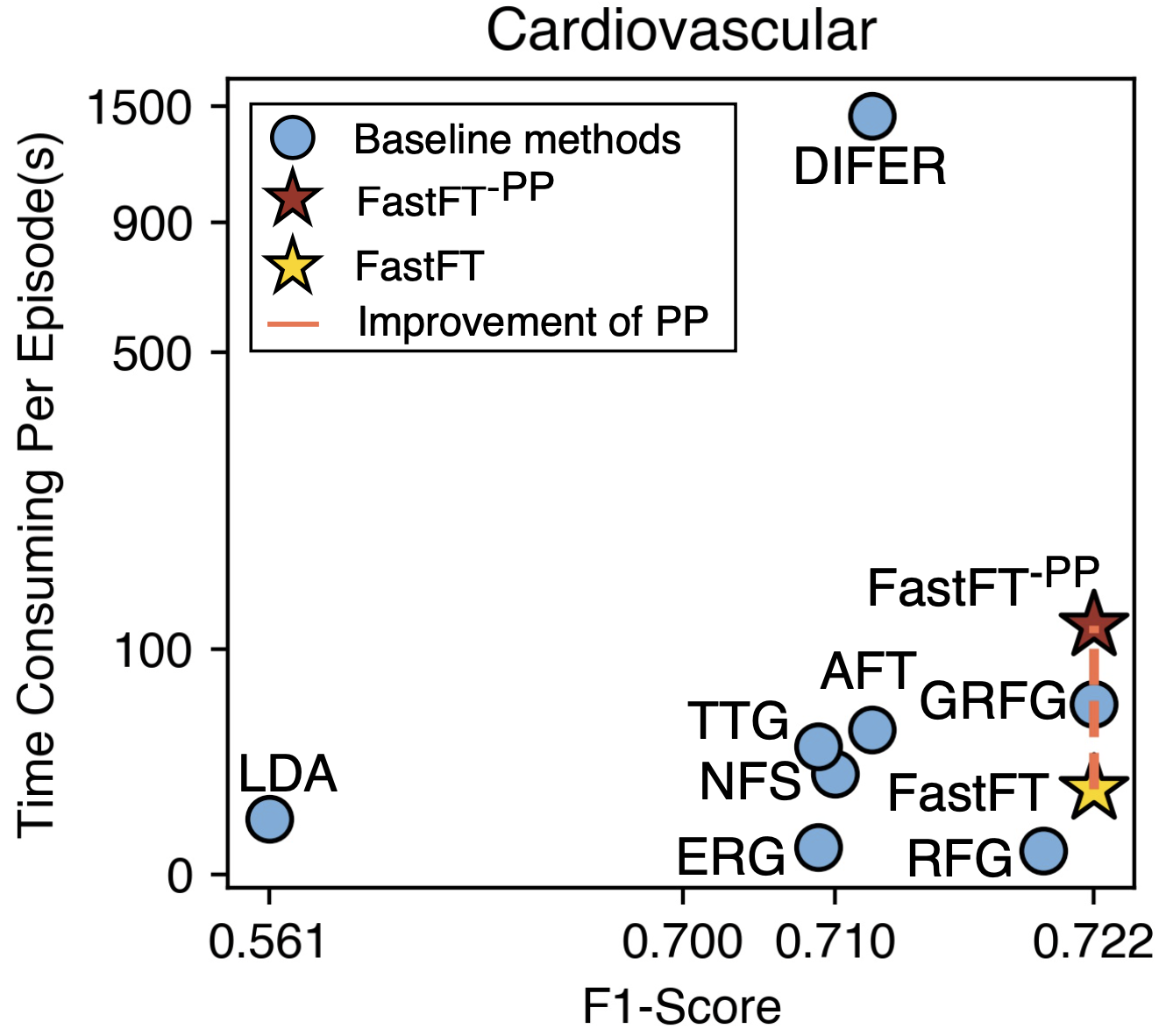}
    \end{minipage}
    \begin{minipage}{0.24\textwidth}
        \centering
        \includegraphics[width=1\linewidth]{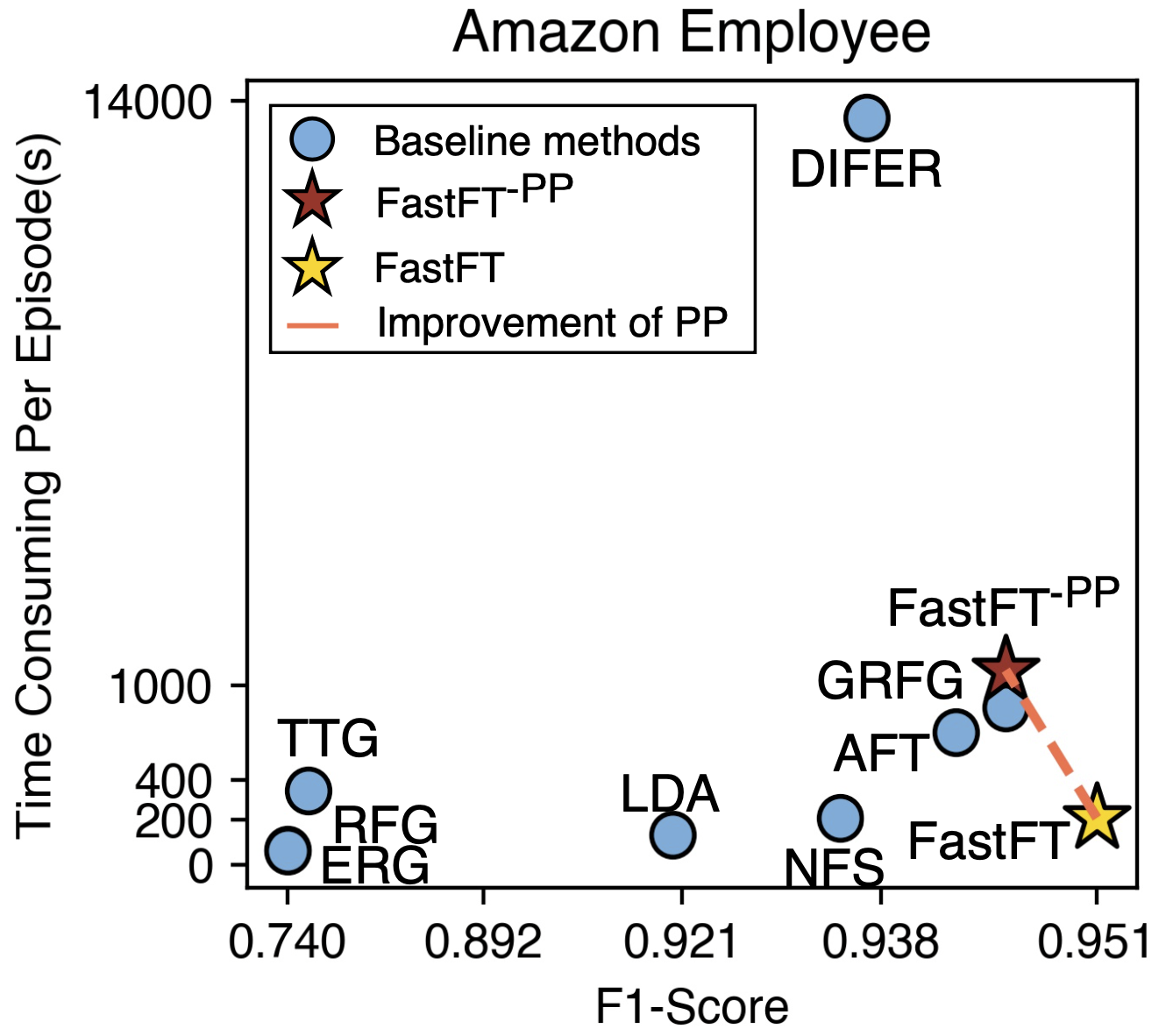}
    \end{minipage}
    \caption{\color{black}Comparison of \model, \(\model^{-PP}\) and baselines in downstream task performance and time consumption. By incorporating an efficient performance predictor, our method explores high-quality features while consuming less time.}
    \vspace{-0.5cm}
    \label{fig:baseline}
\end{figure*}
\noindent\textbf{\textcolor{black}{Impact of Reinforcement Learning Framework.}} 
\textcolor{black}{We replaced the Actor-Critic framework in \model\ with DQN~\cite{dqn}, DDQN~\cite{ddqn}, DuelingDQN~\cite{dueling}, and DuelingDDQN~\cite{dueling}. 
From Figure~\ref{fig:rl_ablation}, the Actor-Critic framework consistently outperforms the other methods while showing a faster convergence.}

\noindent\textbf{\textcolor{black}{Impact of Sequential Modeling Method. }}
\textcolor{black}{We replaced the LSTM model in the Performance Predictor and the Novelty Estimator with Transformer~\cite{vaswani_attention_2017} and RNN~\cite{liu_recurrent_2016}, denoted as \({\model}^{T}\) and \({\model}^{R}\), respectively. 
Figure~\ref{fig:sequence} demonstrates that LSTM performs comparable to RNN and Transformer models, yet with significantly faster runtime. 
The underlying drive could be the complexity of the Feature Transformation Sequence, which does not require relatively sophisticated modeling methods. }

\subsection{Study of the Overall Running Time Analysis}
\label{time.experiment}

This experiment aims to answer the question: \textit{Could \model{} mitigate the time-consuming issue?} 
It is illustrated in Figure~\ref{fig:baseline} that \model\ outperforms all baselines with comparable time-cost to expansion-reduction methods while being significantly faster than iterative-feedback methods and generative-based methods. 
Furthermore, we can observe that \model{} has equivalent performance to \model$^{-PP}$ but only with 20\% of the latter total run-time.
Considering the dataset's scale, as the dataset's size increases, this reduction proportion in time increases as well. 
{\color{black}
Another interesting observation is that in some cases, \model\ could outperform \model$^{-PP}$, while the latter uses downstream task evaluation in each step.
The underlying driver could be the Performance Predictor, which is only trained on the most critical memory. 
It might guide the whole framework to focus more on the high-reward transformation sequence, thus increasing performance.}
These observations show the remarkable scalability of our method.

\subsection{Study of the Model Scalability Check}\label{scalability1}
\begin{figure}[!t]
\centering
\includegraphics[width = 1\linewidth]{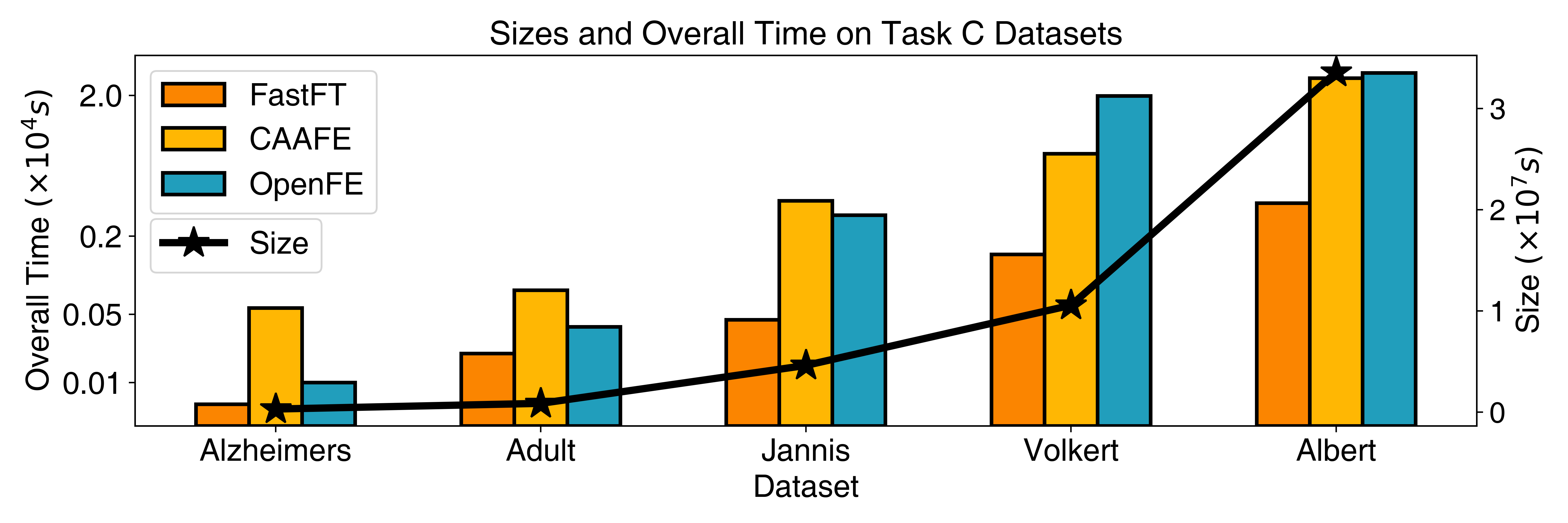}
\caption{\textcolor{black}{Overall model scalability test regarding different dataset size (Features $\times$ Samples).}}
\vspace{-0.5cm}
\label{fig:scalability}
\end{figure}
This experiment aims to answer the question: \textit{Can \model\ maintain reasonable runtime scalability as dataset size increases? }
\textcolor{black}{
We measured the runtime across four large datasets for the classification task.
As illustrated in Figure ~\ref{fig:scalability}, \model\ demonstrates better scalability than baseline methods, CAFFE and OpenFE.
The experimental results highlight significant differences in scalability across the three frameworks. 
\model\ consistently demonstrates lower runtime than OpenFE (from 41.14\% to 7.39\%) and CAAFE (from 7.35\% to 18.80\%), particularly in larger datasets. 
The longer runtime of CAAFE is primarily attributed to the inefficiency of using large language models to process dataset descriptions, which results in substantial time consumption for smaller datasets. 
However, as the dataset size increases, CAAFE's runtime increases slower than OpenFE, yet still significantly outpaces FastFT. This is because OpenFE evaluates each step based on downstream tasks, which imposes a considerable computational bottleneck, particularly for larger datasets.
In contrast, \model\ benefits from using the Performance Predictor and Novelty Estimator, which adaptively replace downstream task evaluations. 
This results in superior scalability compared to OpenFE, as it avoids the computational bottlenecks associated with evaluating the entire generated dataset, leading to more efficient performance even as the dataset scales. 
}

\subsection{Study of the Spatial Complexity Analysis}\label{gpu}
\label{sec:Study of Spatial Complexity}


\begin{figure}[!t]
    \centering
    \subfloat[Seq-length and GPU allocation]{
    \label{fig:gpu1}
    \includegraphics[width=0.5\linewidth]{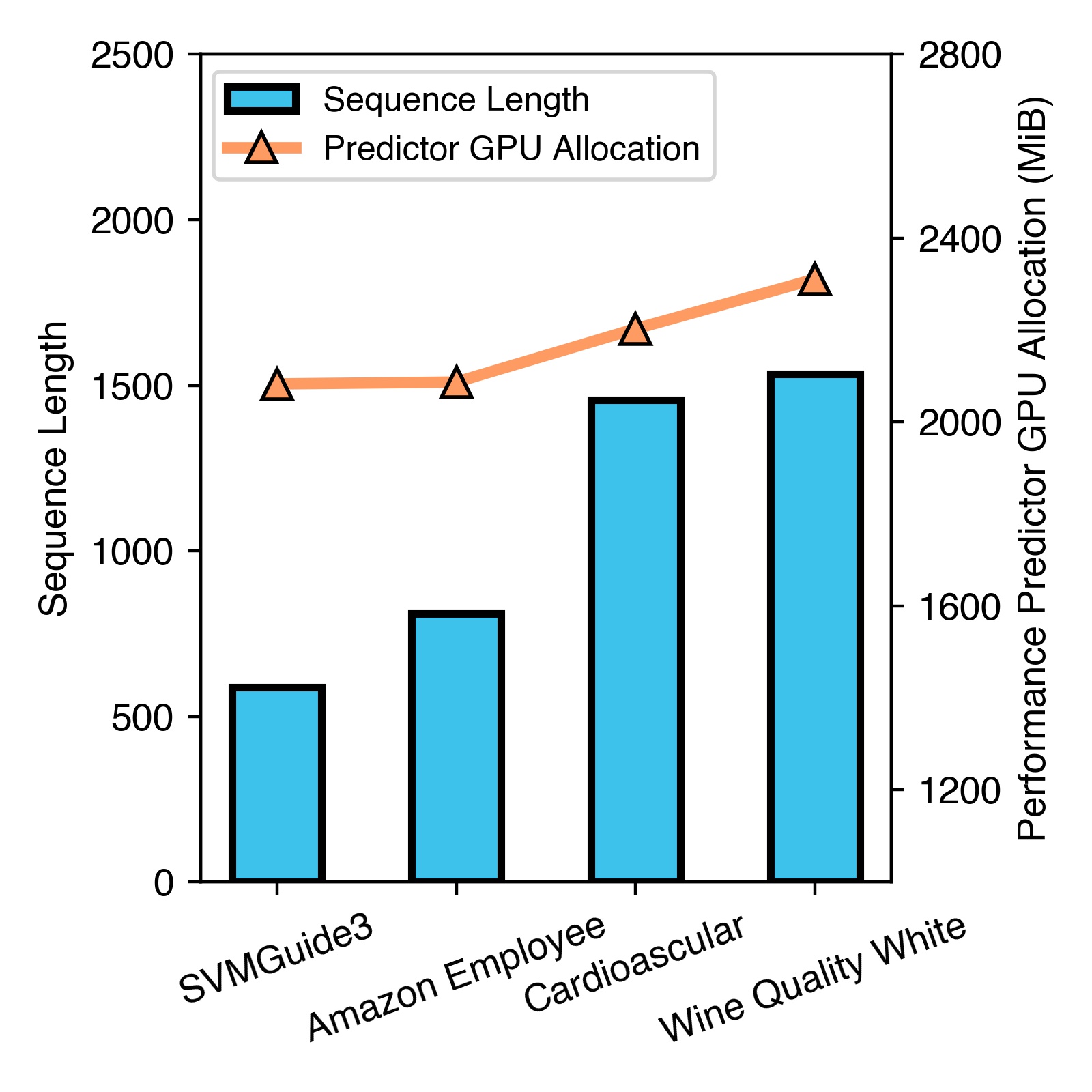}
    }
    \subfloat[GPU allocation and time]{
    \label{fig:gpu2}
    \includegraphics[width=0.5\linewidth]{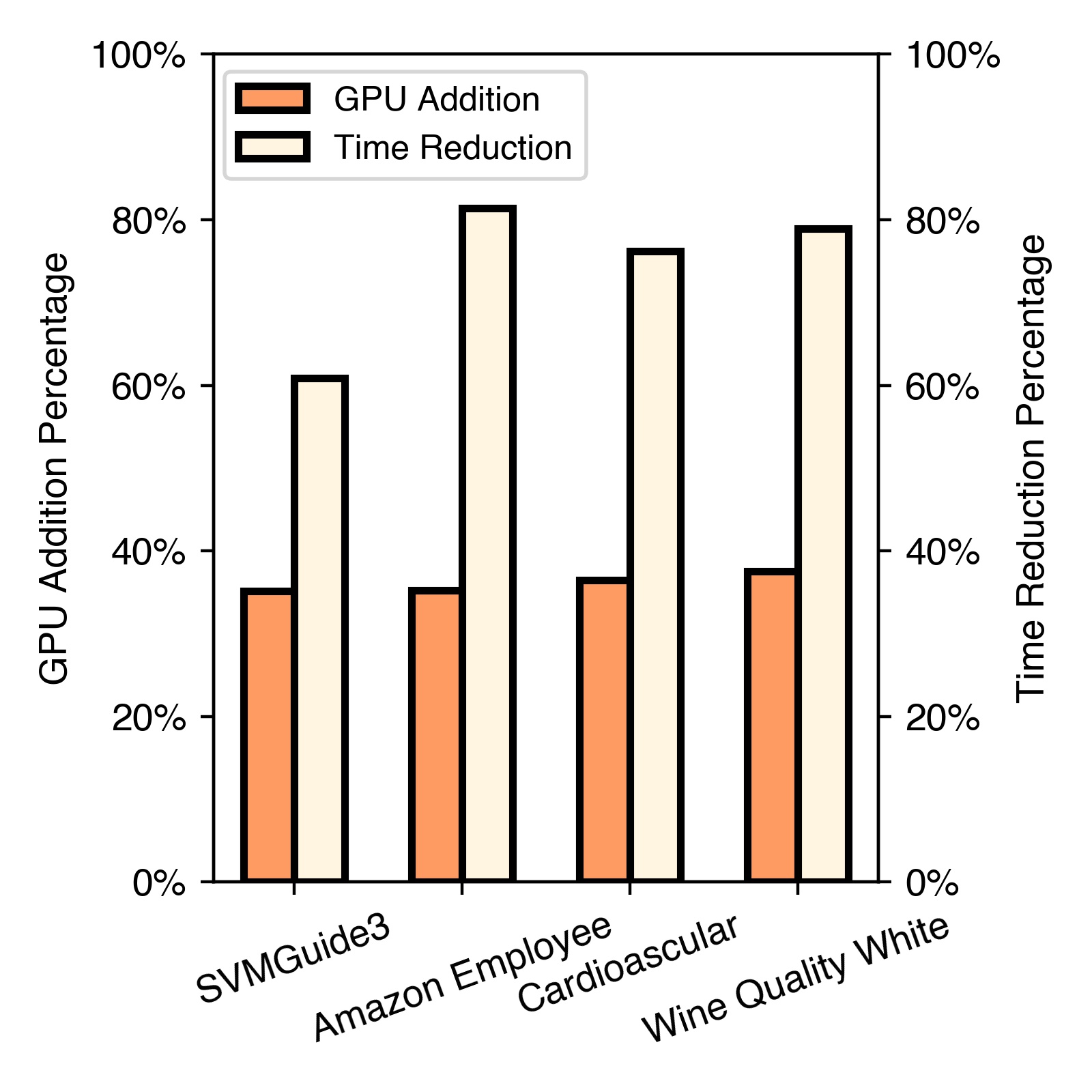}
    }    
    \caption{(a) Sequence length and predictor GPU allocation. (b) The trade-off between GPU allocation and time consumption.}
    \label{fig:gpu}
\end{figure}


\begin{figure*}[!t]
    \centering
    \subfloat[\textcolor{black}{$\alpha$ (Ajust NE Threshold)}]{
    \label{fig:threshold1}
    \begin{minipage}{0.24\linewidth}
        \centering
        \includegraphics[width=\linewidth]{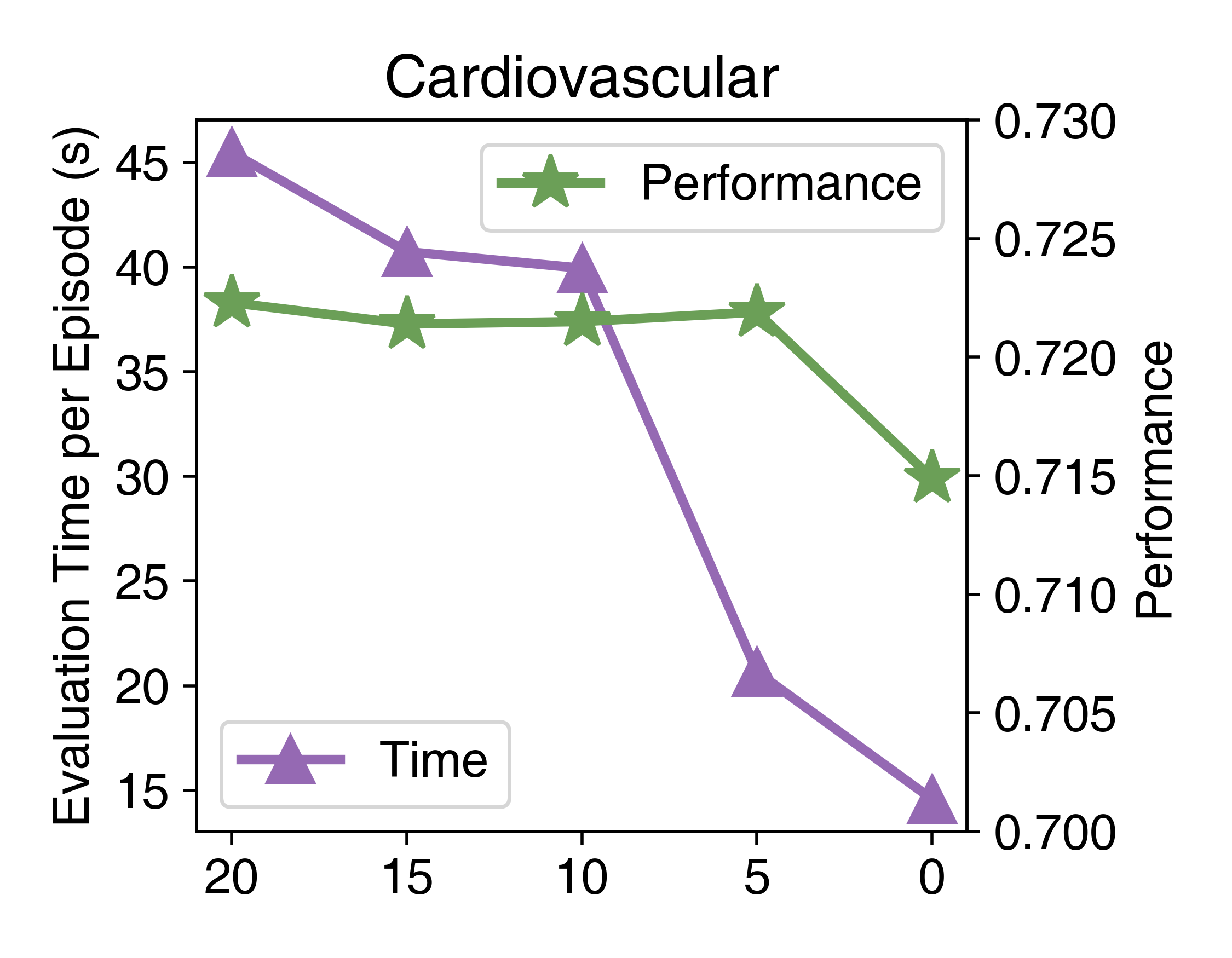}
    \end{minipage}
    \begin{minipage}{0.24\linewidth}
        \centering
        \includegraphics[width=\linewidth]{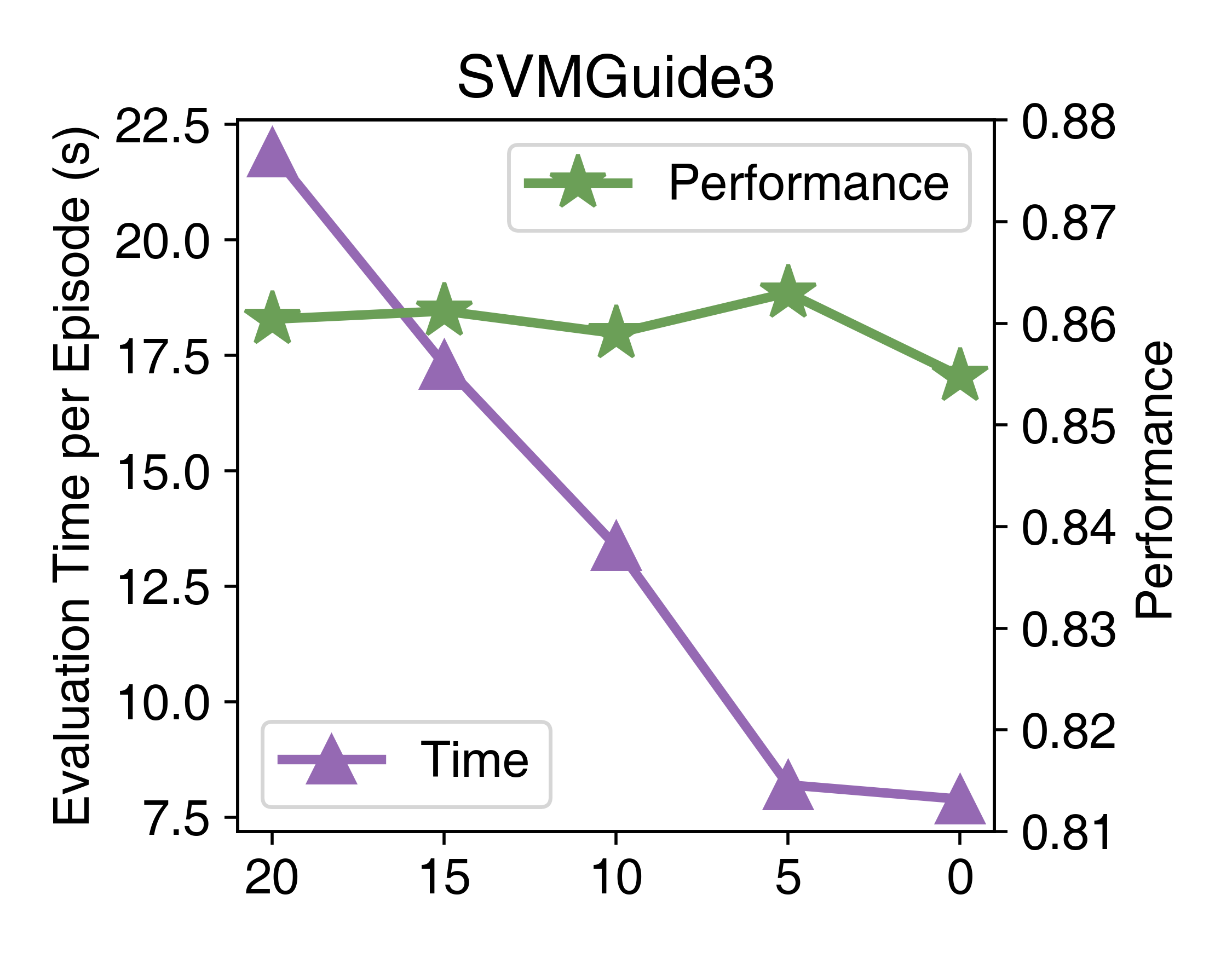}
    \end{minipage}
    }
    \subfloat[\textcolor{black}{$\beta$ (Ajust PP Threshold)}]{
    \label{fig:threshold2}
    \begin{minipage}{0.24\linewidth}
        \centering
        \includegraphics[width=\linewidth]{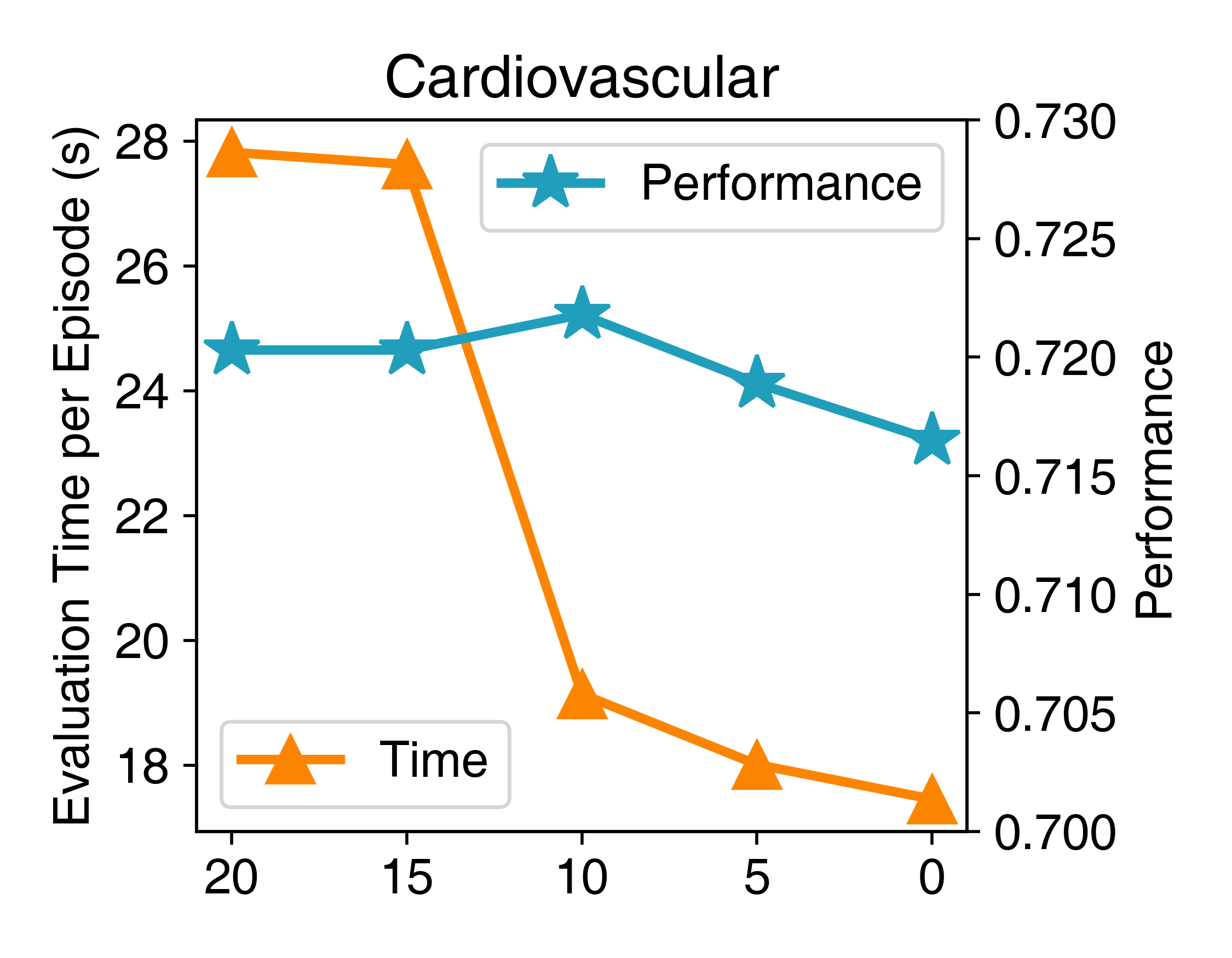}
    \end{minipage}
    \begin{minipage}{0.24\linewidth}
        \centering
        \includegraphics[width=\linewidth]{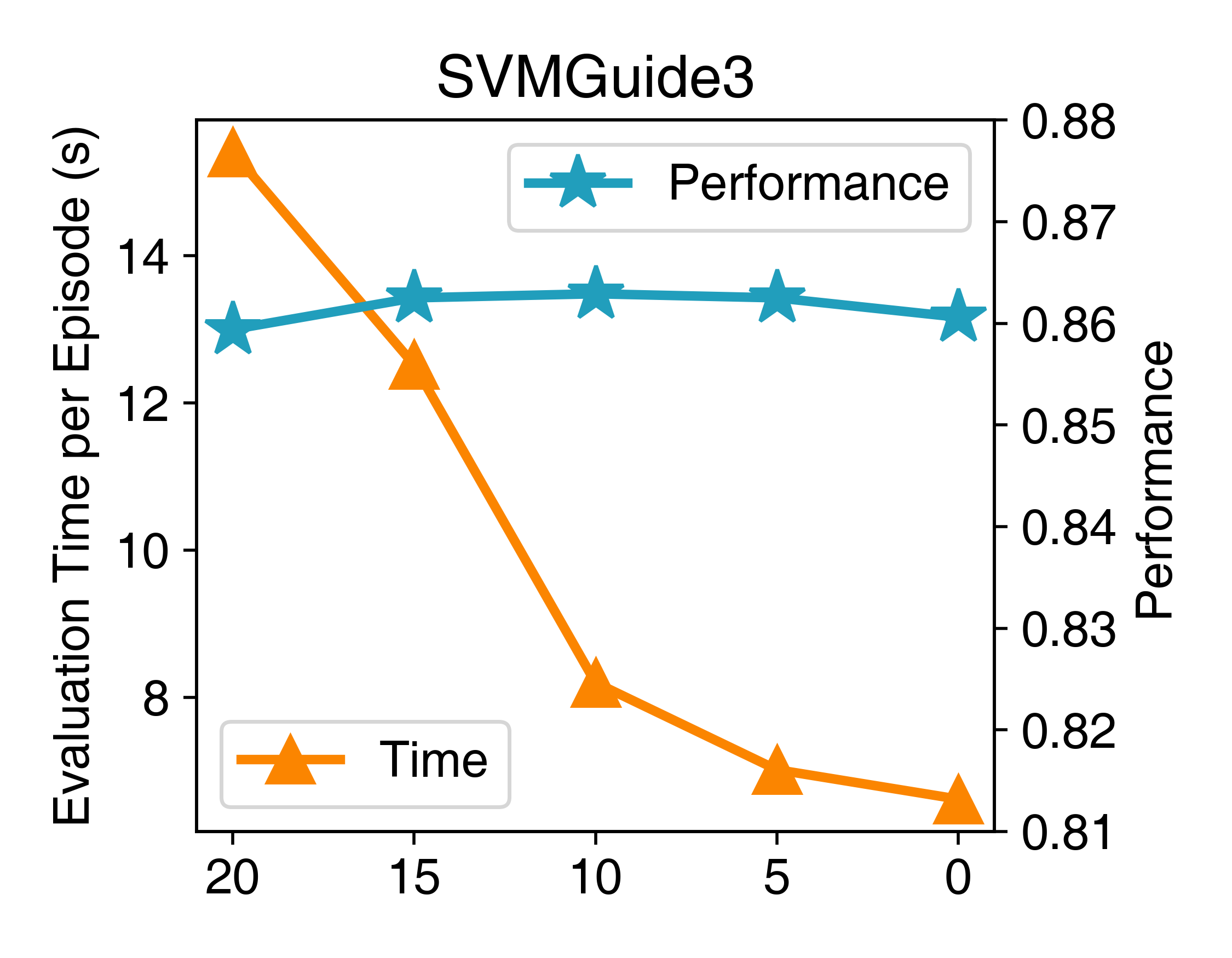}
    \end{minipage}
    }    
    \caption{\textcolor{black}{The impact of (a) performance predictor threshold $\alpha$ and (b) novelty estimator threshold $\beta$ on time consuming and performance. With higher $\alpha$ and $\beta$, the evaluation time significantly reduces, while performance exhibits minor fluctuations.}}
    \vspace{-0.4cm}
    \label{fig:threshold}
\end{figure*}
\begin{figure*}[!t]
    \centering
    \label{fig:hyper_ws}
\includegraphics[width=0.24\linewidth]{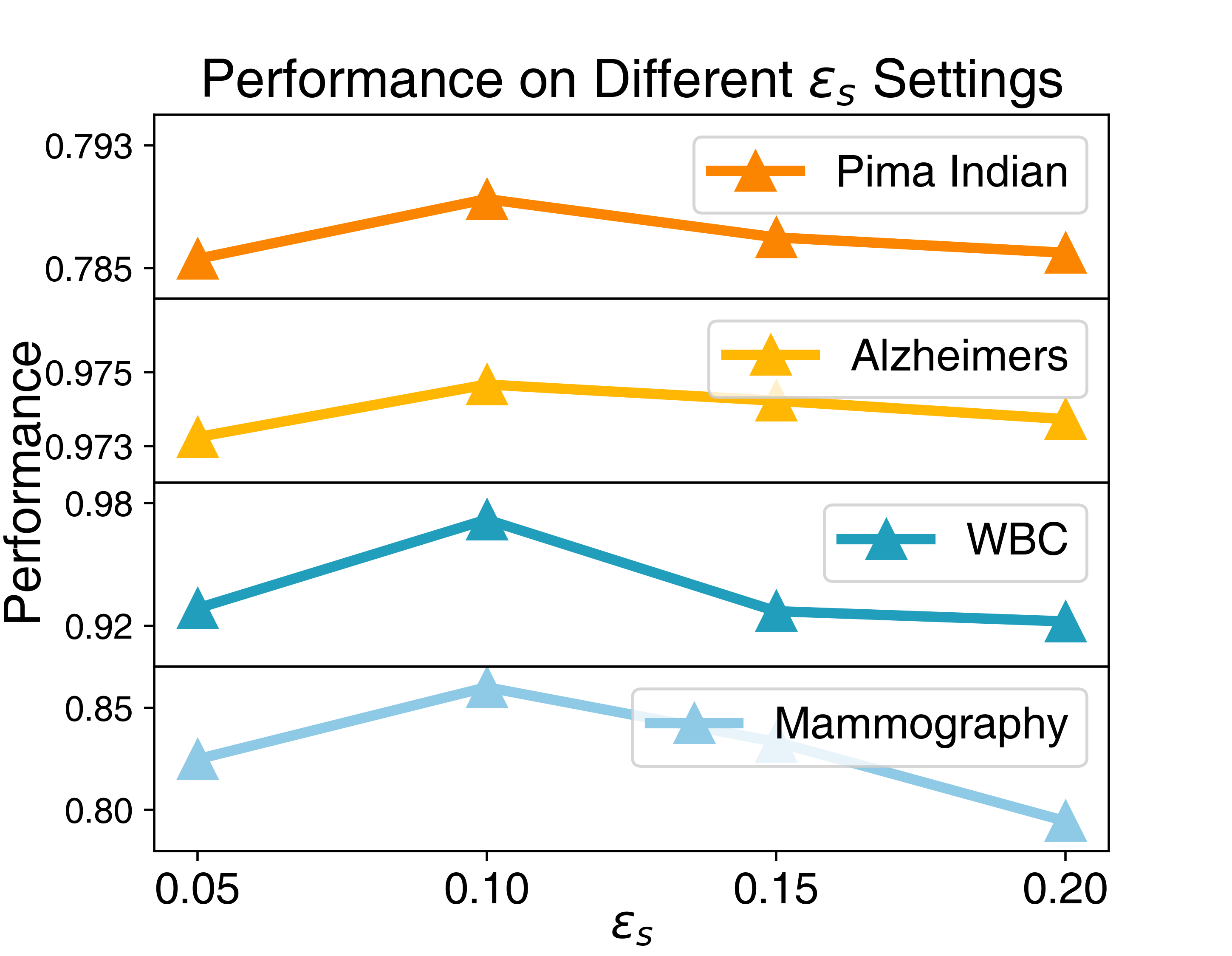}\hspace{-0.1cm}
    \label{fig:hyper_we}
\includegraphics[width=0.24\linewidth]{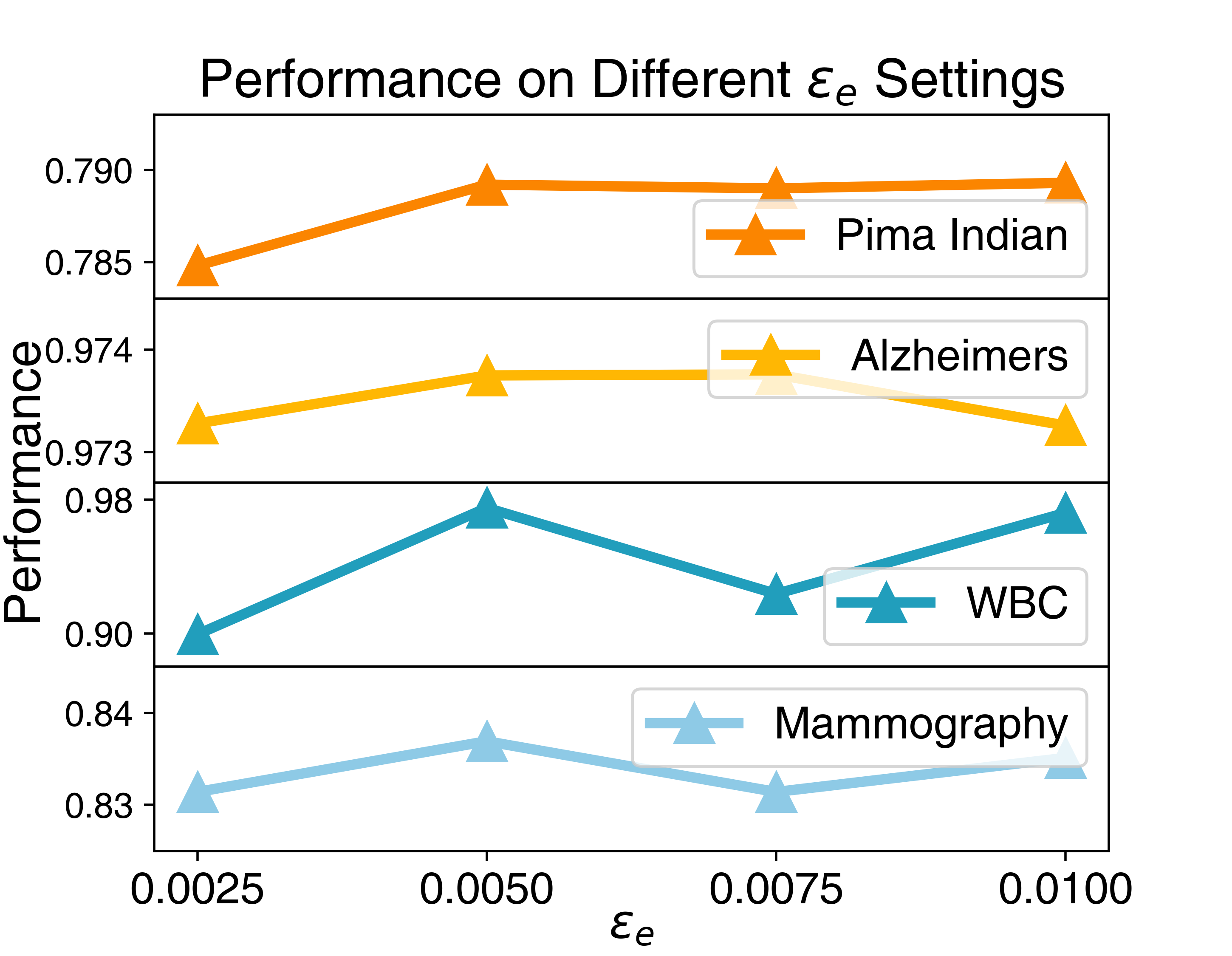}\hspace{-0.1cm}
    \label{fig:hyper_steps}
\includegraphics[width=0.24\linewidth]{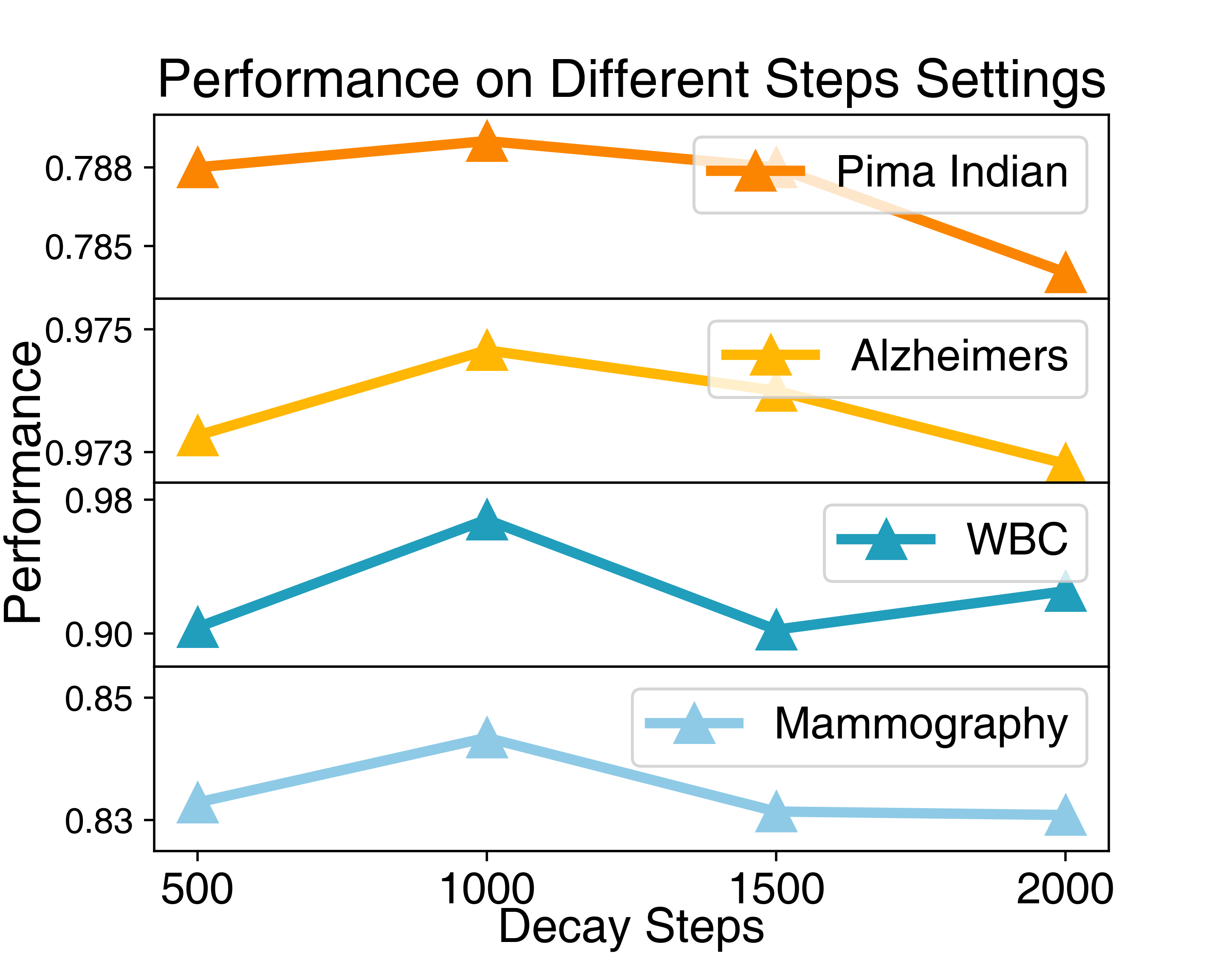}\hspace{-0.1cm}
    \label{fig:hyper_memory}
\includegraphics[width=0.24\linewidth]
    {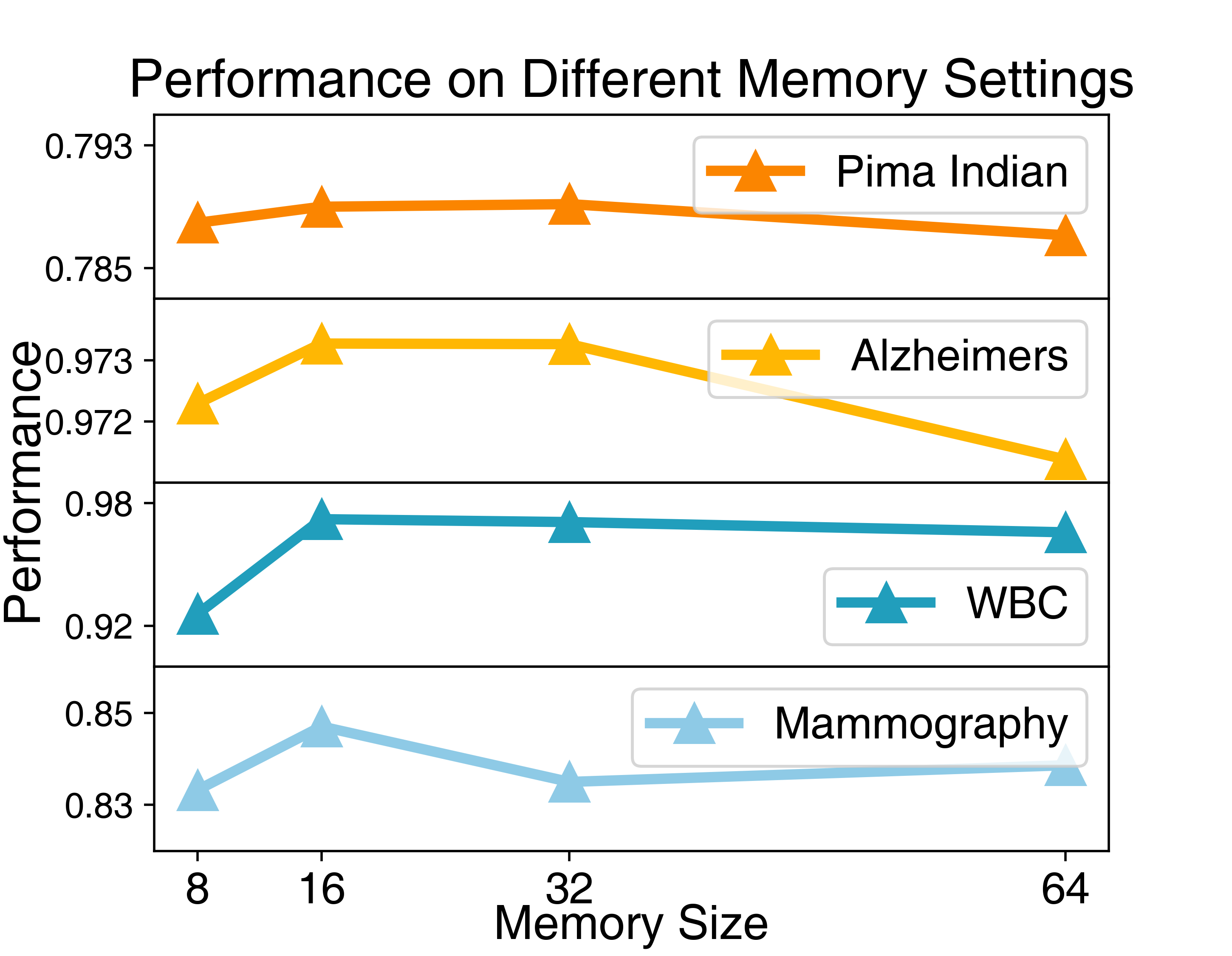}
    \caption{\textcolor{black}{Hyperparameter studies on different novelty reward weight and memory size settings for four Datasets.}}
    \vspace{-0.4cm}
    \label{fig:hyper}
\end{figure*}
This experiment aims to answer the question: \textit{What is the GPU resource utilization of predictor model?}
Figure~\ref{fig:gpu1} illustrates the variation in GPU usage of the Performance Predictor with changes in sequence length. 
GPU resource consumption increases slowly with longer sequences. 
This is due to the recurrent architecture of the Performance Predictor, which makes it less sensitive to the sequence length.
Figure~\ref{fig:gpu2} shows the trade-off between the additional GPU consumption introduced by the Performance Predictor and the reduction in time consumption.  
We observe that while GPU usage slightly increases, our method's time consumption significantly decreases.
The underlying reason is that we use a forward network to replace the dense and complex downstream task computation, leading to substantial time savings while only requiring acceptable GPU resources.
In summary, this experiment demonstrates the GPU utilization of \model{}.

\subsection{Study of the Hyperparameter}\label{hyper}

\noindent\textbf{Efficiency-Efficacy Trade-off ($\alpha$, $\beta$).}
We modify the setting of two hyperparameters, $\alpha$ and $\beta$, in Section~\ref{opt}. 
In detail, an increased value of $\alpha$ results in more downstream assessment.
Conversely, a higher $\beta$ means that more transformation sequences with lower novelty are evaluated downstream. 
When these two values are set to 0, the evaluation component will take over all exploration and optimization processes. 
To exam the $\alpha$, we fix $\beta$ at 5 and varied $\alpha$ from 0 to 20.
For $\beta$, we fix $\alpha$ at 10 and varied $\beta$ from 0 to 20.
As illustrated in Figure~\ref{fig:threshold}, with $\alpha$ or $\beta$ decreases, evaluation time significantly reduces, while performance exhibits only minor fluctuations except when $\alpha$ and $\beta$ are set to 0.
Regarding the reduced time consumption, the underlying drive is that the decrease in $\alpha$ or $\beta$ minimizes the proportion of downstream tasks takeover, thus reducing the time cost. 
Nevertheless, when threshold parameters are reduced to 0, downstream tasks do not evaluate any of the feature sets. 
This setting will prevent reinforced agents from receiving accurate performance feedback, leading to potential degeneration during exploration.
According to this hyperparameter study, we set the $\alpha$ and $\beta$ as a reasonable constant.

\noindent\textbf{\textcolor{black}{Novelty Reward Weight ($\epsilon_s$, $\epsilon_e$), Decay Steps($M$) and Memory Size ($S$).}}
\textcolor{black}{
From Figure~\ref{fig:hyper}, we observed that the model’s performance remains relatively stable across a range of different settings, showcasing the model's strong ability to generalize to hyperparameter choices. 
Besides, when applied to four diverse datasets of varying sizes and tasks, the performance trends exhibit consistency despite variations in hyperparameter settings. 
This highlights that the optimal hyperparameter settings remain effective regardless of the underlying task or dataset scale. 
Additionally, the results show that the memory size hyperparameter does not benefit from being arbitrarily large. 
Smaller memory sizes contribute to more effective performance, ensuring that key memories are updated. 
This is particularly advantageous for more complex datasets, such as Alzheimers and Mammography, where timely updates to the model's strategy and fine-tuning of the Evaluation Component are critical. 
Based on these observations, we set the key hyperparameters as follows: $\epsilon_s=0.10$, $\epsilon_e=0.005$, $M=1000$, and $S=16$. }

\begin{table}[!t]
\centering
\caption{Robustness check of \model\ with distinct ML models on German Credit dataset in terms of F1-Score.}
\label{table:robust}
\centering
\medskip
\resizebox{\linewidth}{!}{
\begin{tabular}{lcccccc}
\toprule 
      & RFC   & XGBC  & LR    & SVM-C & Ridge-C & DT-C\\ \midrule
ATF   & 0.751 & 0.714 & 0.671 & 0.672 & 0.658   & 0.692\\ 
ERG   & 0.661 & 0.741 & 0.757 & 0.691 & \textbf{0.758}   & 0.690\\
LDA   & 0.627 & 0.650 & 0.576 & 0.574 & 0.574   & 0.636\\
NFS   & 0.765 & 0.741 & 0.761 & 0.759 & 0.752   & 0.669\\
RDG   & 0.751 & 0.741 & 0.760 & 0.758 & 0.750   & 0.664\\
TTG   & 0.731 & 0.746 & 0.751 & 0.750 & 0.744   & 0.673\\ 
GRFG  & 0.763 & 0.747 & 0.755 & 0.753 & 0.744   & 0.689\\
DIFER & 0.752 & 0.741 & 0.576 & 0.640 & 0.719   & 0.664\\ 
\textbf{\model\ } & \textbf{0.777} & \textbf{0.750} & \textbf{0.762} & \textbf{0.763} & \textbf{0.758} & \textbf{0.695} \\ \bottomrule
\end{tabular}}
\end{table}
\begin{figure}[!ht]
    \centering
    \subfloat[\color{black}Novelty Distance]{%
        \includegraphics[width=0.23\textwidth]{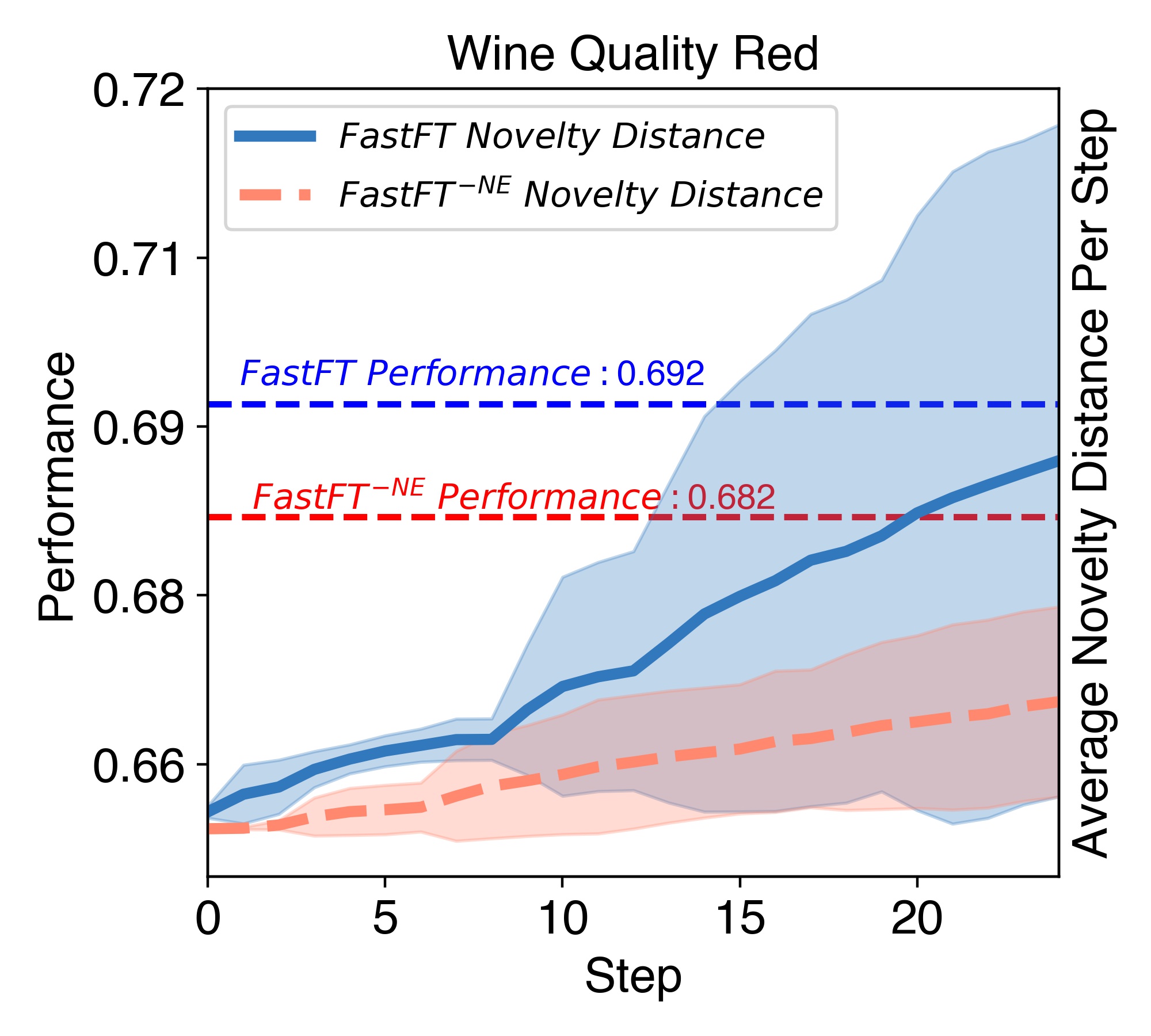}%
    }
    \hfill
    \subfloat[\color{black}Performance and Combinations.]{%
        \includegraphics[width=0.25\textwidth]{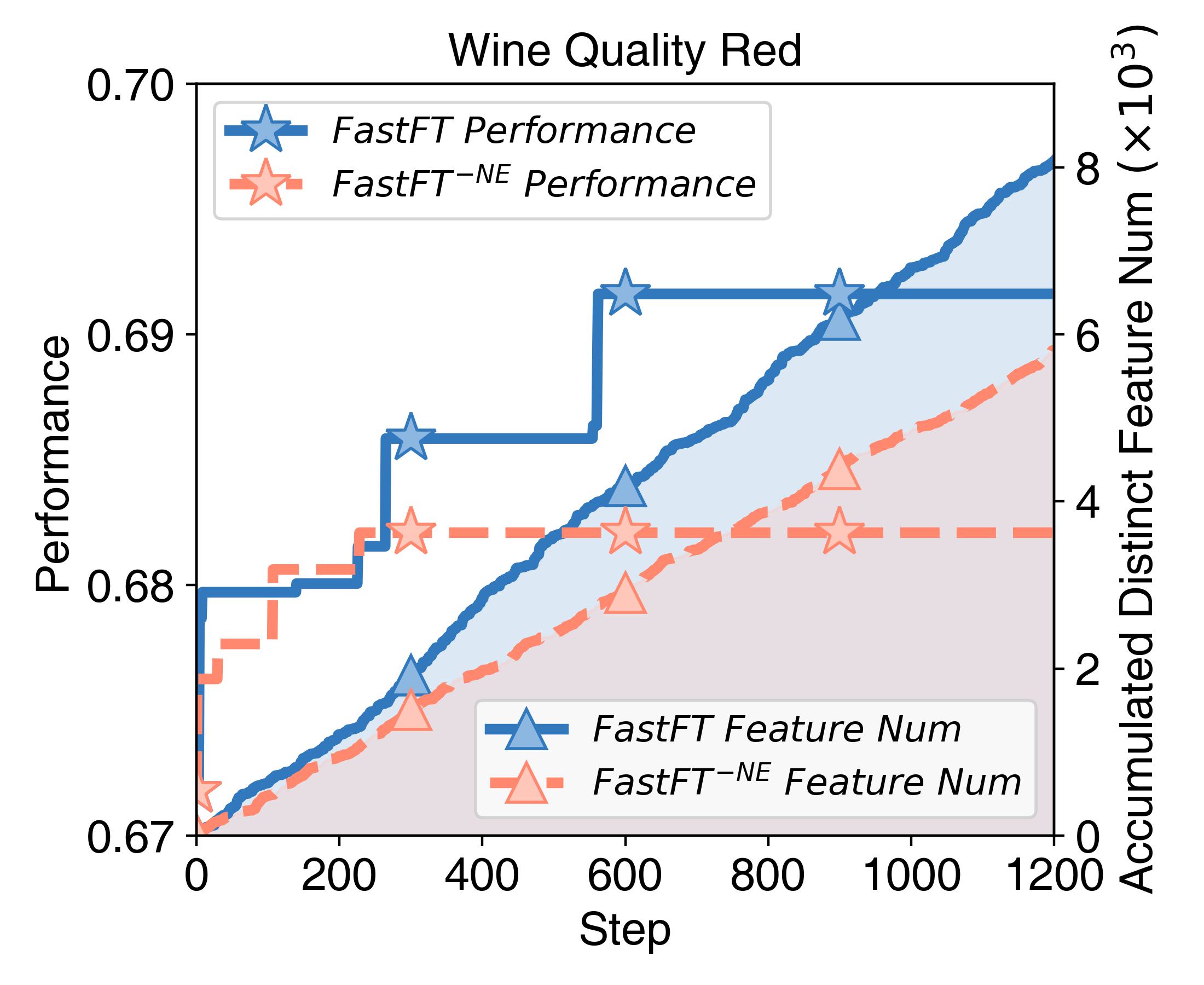}%
    }
    \caption{The comparison between \model\ and \model$^{-NE}$ in terms of average novelty distance of generated features, the unencountered feature number, and its corresponding downstream ML task performance.}
    \vspace{-0.5cm}
    \label{fig:novelty}
\end{figure}


\begin{figure*}[!t]
\centering
\includegraphics[width = 0.8\linewidth]{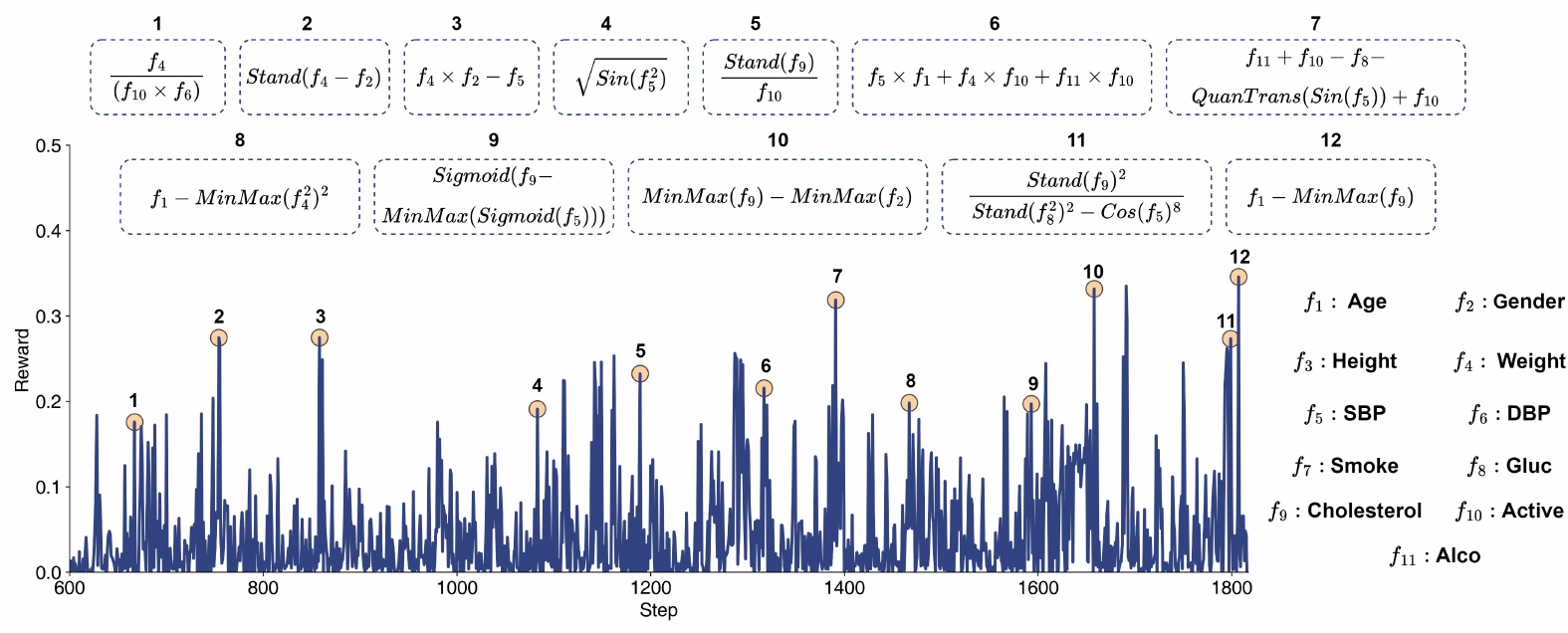}
\caption{Distinct features associated with peaks in the reward function. 
These features not only improve the dataset's performance on downstream tasks but also reveal latent knowledge within the data.}
\label{fig:case}
\vspace{-0.4cm}
\end{figure*}

\subsection{Study of the Robustness Check}\label{robust}

This experiment aims to answer the question: \textit{Are our generative features robust across various machine learning models employed in downstream tasks?} 
We conducted the experiment using the German Credit dataset for classification tasks and applied a range of feature transformation baseline techniques. 
We assessed the robustness using a Random Forest Classifier (RFC), XGBoost Classifier (XGBC), Logistic Regression (LR), SVM Classifier (SVM-C), Ridge Classifier (Ridge-C), and Decision Tree Classifier (DT-C).
The experiments were evaluated in terms of F1-Score. 
The results presented in Table~\ref{table:robust}, demonstrate that the features generated by \model\ consistently outperform those produced by other techniques on various downstream ML tasks. 
The reason is that the integrated reward mechanism guides the cascading agents to explore a broader and more efficient feature space, thereby achieving a globally optimal feature set and indicating superior robustness.

\subsection{Study of the Novelty Reward}\label{novelty}
This experiment aims to answer the question: \textit{What is the impact of the Novelty Reward?}
To answer the question, we design a metric, novelty distance, to represent the distinction of generated feature combinations.
This metric is defined as the minimum cosine distance between the current and all collected historical feature set embedding. 
In Figure~\ref{fig:novelty}, we illustrate the accumulated average novelty distance of generated features, the number of unencountered feature combinations, and the corresponding performance between \model\ and \({\model}^{-NE}\) at each step. 
The first finding is that \model\ attains a higher average novelty distance for each exploration step (in Figure~\ref{fig:novelty} (a)), meanwhile generating more unencountered feature combinations (as shown in Figure~\ref{fig:novelty} (b)). 
The potential cause of these phenomena is that the novelty reward enables the agent to expand the search space, leading to the discovery of more unique feature combinations.
Furthermore, the result suggests that features with higher novelty tend to perform better in downstream tasks. 
The reason is that the novelty reward, in collaboration with the performance reward, drives the agent to search for a global optimum, thereby enhancing performance. 
In summary, this experiment underscores the significant role of novelty reward in discovering feature combinations and enhancing performance.

\section{Case Studies}
\begin{table}[!t]
\centering
\caption{Top-10 important features on original and transformed Wine Quality Red datasets. \model\ generates more important features and exhibits strong traceability.}
\resizebox{\linewidth}{!}{
\begin{tabular}{cccc}
\toprule
\multicolumn{2}{c}{\textbf{Wine Quality Red}} & \multicolumn{2}{c}{\textbf{\model}}     \\
\multicolumn{1}{c}{Feature} & Importance & Feature & Importance \\ \midrule
$f_{11}:$ alcohol              & 0.150 & $(f_{3} \times f_{9} + 1) \times f_{4} + f_{1} \times f_{9}$ & 0.026  \\
$f_7:$ sulfur dioxide & 0.110 & $f_{1} \times f_{9} + 1) \times f_{4}+ f_{3} \times f_{9}$  & 0.022   \\
$f_{10}:$ sulphates         & 0.108 & $f_{11}$ & 0.020   \\
$f_2:$ volatile acidity     & 0.097 & $f_{4} + {f_{3}}^{2} \times {f_{9}}^{2}$ & 0.019    \\
$f_8:$ density              & 0.085 & $f_{7} \times f_{9} \times f_{4}+ f_{3} \times f_{9}$ & 0.019   \\
$f_5:$ chlorides            & 0.081 & $f_{4} + f_{3} \times{f_{9}}^{2} \times f_{7}$ & 0.018   \\
$f_1:$ fixed acidity        & 0.078 & $f_{3} \times f_{9} \times (f_{4} + 1)$ & 0.017   \\
$f_9:$ pH                   & 0.077 & $f_{4} + f_{3} \times f_{9} \times f_{7}$ &  0.016  \\
$f_3:$ citric acid          & 0.073 & $f_{4} + f_{3} \times f_{9} \times (f_{4} + 1)$ & 0.015   \\
$f_4:$ residual sugar       & 0.071 & $f_{2}$ & 0.015   \\
\midrule
\multicolumn{1}{c}{F1-Score: 0.672} & Sum: 0.931 & F1-Score: 0.695 & Sum: 0.188 \\ \bottomrule
\end{tabular}
\vspace{-0.8cm}}
\label{case2}
\end{table}

\subsection{Case Study on Generated Features}\label{case_study2}
We analyzed the top 10 most important features among the generated feature set. 
As illustrated in Table~\ref{case2}, the top 10 features produced by our approach show a more balanced importance score. 
This is due to \model\ generating a higher number of features and replacing useless features, thus showing a balanced distribution of the importance and improving performance in downstream tasks. 
Additionally, our method is capable of explicitly outlining the feature transformations, making the transformation process transparent. 
These traceable attributes can aid experts in uncovering new domain mechanisms, especially in AI4Science domain.

\subsection{Case Study on Feature Transformation Process}\label{case_study1}
This experiment was conducted on the Cardiovascular dataset, which predicts cardiovascular disease risk based on personal lifestyle factors and medical indicators.
We examined the steps at which the reinforcement learning reward peaked and identified distinct features generated at these steps. 
The results are illustrated in Figure ~\ref{fig:case}. 
Our method demonstrates the ability to produce traceable features, establishing a clear mathematical relationship between the original and generated features. This transparency facilitates the analysis of their significance and the discovery of hidden knowledge. 
For instance, the new feature $Weight / (Active \times\ DBP)$ is generated at the step marked by point 1. 
In particular, $Active$ represents the level of physical activity, and DBP refers to Diastolic Blood Pressure. 
Generally, DBP tends to increase with weight and decrease with higher physical activity\cite{staessen1988relationship,borjesson2016physical}. 
This generated feature highlights DBP values that deviate from this expected pattern, suggesting abnormalities relative to weight and physical activity levels. 
Such deviations may have potential utility in the diagnosis of cardiovascular disease. 

In summary, those generated features with high novelty are traceability, which could potentially reveal the hidden knowledge within the dataset. 


\section{Related Work}
Feature transformation refers to the process of generating high-quality features by applying a series of mathematical transformations to the original features.
High-quality datasets suit the needs of machine learning algorithms better~\cite{chen2021techniques,zha2023data}.
Automated feature transformation implies that the machine performs this task automatically without requiring prior knowledge or intervention from humans~\cite{lam2017one,zhang2023openfe}.
There are four mainstream approaches:
(1) \textit{expansion-reduction} based method~\cite{kanter2015deep,horn2019autofeat,khurana2016cognito,lam2017one,khurana2016automating} randomly selects features to transform.
Due to its inherent randomness, this method is unstable and has a limited exploration space, resulting in suboptimal performance~\cite{katz2016explorekit,dor2012strengthening}.
(2) \textit{iteratively-feedback} based models~\cite{khurana2018feature,tran2016genetic,kdd2022,xiao2022traceable,zhu2022evolutionary,xiao2024traceable} integrate feature selection and feature transformation into a single learning process, typically optimizing through evolutionary algorithms or reinforcement learning~\cite{ren2023mafsids}.
However, this method suffers from time bottlenecks due to repeatedly using inefficient downstream tasks as feedback~\cite{huang2024enhancing}.
(3) \textit{AutoML} based method~\cite{wang2021autods,zhu2022difer,xiao2023discrete,ying2023self} models AFT task as a continuous space optimization problem and achieves remarkable performance.
However, this method is limited by the quality of the collected transformation data, which can lead to instability or poor traceability during the generation phase.
(4) \textit{Large language model} based approaches~\cite{zhang2024dynamic,hollmann2023large} attempt to leverage the semantic information of feature names for transformation, generating high-quality features. 
However, since feature names are often obscured in real datasets, the applicability of such methods remains limited.
To overcome these problems, \model\ integrates empirical-based performance prediction with orthogonally initialized neural networks and a revisitation mechanism, improving search productivity and framework performance across various domains.

\section{Limitations and Future Work}
{\color{black}
\noindent\textbf{Capabilities of LLMs in Feature Transformation.}
LLMs have not been incorporated into our approach due to the following four challenges: 
(1) Missing Semantic Information: LLMs typically rely on dataset descriptions and feature names, which are often absent, thus limiting their effectiveness; 
(2) Hallucination: LLMs may generate irrelevant or unsupported content~\cite{hassan2023chatgpt,qin2025scihorizon}, leading to misleading features. 
(3) Limitations in scalability: In our experiments, frameworks such as CAAFE could not handle large-scale datasets.
(4) Insufficient generalization capabilities: LLM-based method remains limited generalization even trained on extensive dataset collection~\cite{wen2024supervised,cai2023resolving}. 
In future work, we plan to incorporate LLMs from different perspectives. 
For instance, we will enhance the generated feature's interpretability and extract knowledge from unstructured texts, such as research papers, thereby guiding the transformation process to uncover more meaningful features.

\noindent\textbf{Limiation on High-dimension Data and Noisy Data.}
According to our experimental results, feature transformation appears more suitable when the data volume is insufficient to support model convergence.
As the data size increases, deep neural networks can learn more generalized patterns, making feature transformation less significant. 
Besides, our future work will explore integrating noise-robust training strategies to enhance the framework's adaptability in these complex settings. 

}
\section{Conclusion Remarks}

{\color{black}In this paper, we propose \model, a framework for efficient feature transformation in data-centric machine learning pipelines. By decoupling feature evaluation from downstream task performance through a Performance Predictor, \model\ reduces reliance on time-consuming evaluations over large datasets, addressing runtime bottlenecks in feature transformation. To tackle the sparse rewards in reinforcement learning, we introduce a novelty estimation method that encourages exploration of diverse feature combinations. Incorporating both novelty and performance into a prioritized memory buffer ensures effective revisiting of critical transformations, boosting learning efficiency and convergence. \model\ optimizes feature transformation processes, crucial for preparing high-quality datasets. In future work, we aim to refine the evaluation components to enhance their accuracy and generalization on larger, more complex datasets.}

\clearpage
\balance
\normalem
\bibliographystyle{IEEEtran}
\bibliography{sample-base}

\end{document}